%% file: main_paper.tex
\documentclass[lettersize,journal]{IEEEtran}
\usepackage[noend]{algpseudocode}
\usepackage{algorithmicx,algorithm}
\usepackage{cite}
\usepackage{tikz}
\usepackage{amsmath}
\usepackage{comment}
\usepackage{color}
\usepackage{booktabs}
\usepackage{xspace}
\usepackage{amssymb}
\usepackage{amsfonts}           
\usepackage{mathrsfs} 
\usepackage{subfigure}
\usepackage{multirow}
\usepackage{booktabs}
\usepackage{graphicx}
\usepackage{enumitem}
\usepackage{setspace}

\usepackage{hyperref}
\hypersetup{hidelinks} 

\newcommand{\mymodel}{Generative Memory-Guided Semantic Reasoning Model\xspace}
\newcommand{\peicomment}[1]{\textcolor[rgb]{1,0,0} {#1}}

\begin{document}

\title{Generative Memory-Guided Semantic Reasoning Model for Image Inpainting}

\author{Xin Feng,
        Wenjie Pei,
        Fengjun Li,
        Fanglin Chen, ~\IEEEmembership{Member,~IEEE},
        David Zhang, ~\IEEEmembership{Life~Fellow,~IEEE} 
        and~Guangming Lu,~\IEEEmembership{Member,~IEEE}
\thanks{Xin Feng, Wenjie Pei, Fengjun Li, Fanglin Chen and Guangming Lu are with the Department of
Computer Science, Harbin Institute of Technology at Shenzhen, Shenzhen
518057, China (e-mail: fengx\text{\_hit}@outlook.com; wenjiecoder@outlook.com; 20s151173@stu.hit.edu.cn; chenfanglin@hit.edu.cn; luguangm@hit.edu.cn).}%
\thanks{David Zhang is with the School of Science and Engineering, The Chinese University of Hong Kong at Shenzhen, Shenzhen 518172, China (e-mail: davidzhang@cuhk.edu.cn)}
\thanks{Manuscript received Xxxx xx, xxxx; revised Xxxx xx, xxx.}}



\maketitle

\begin{abstract}
\input{0.abstract}
\end{abstract}

\section{Introduction}
\input{1.introduction}

\section{Related Work}
\input{2.Relatedwork}

\section{\mymodel}
\input{3.Method}

\section{Experiments}
\input{4.Experiments}

\section{Conclusion}
\input{5.Conclusion}
\bibliographystyle{IEEEtran}
\bibliography{mybibfile}

\newpage

 




\vfill

\end{document}

%% file: 0.abstract.tex
The critical challenge of single image inpainting stems from accurate semantic inference via limited information while maintaining image quality.
Typical methods for semantic image inpainting train an encoder-decoder network by learning a one-to-one mapping from the corrupted image to the inpainted version.
While such methods perform well on images with small corrupted regions, it is challenging for these methods to deal with images with large corrupted area due to two potential limitations.
1) Such one-to-one mapping paradigm tends to overfit each single training pair of images;
2) The inter-image prior knowledge about the general distribution patterns of visual semantics, which can be transferred across images sharing similar semantics, is not explicitly exploited. 
In this paper, we propose the \textbf{G}enerative \textbf{M}emory-guided \textbf{S}emantic \textbf{R}easoning \textbf{M}odel (\emph{GM-SRM}),
which infers the content of corrupted regions based on not only the known regions of the corrupted image,
but also the learned inter-image reasoning priors characterizing the generalizable semantic distribution patterns between similar images.  
In particular, the proposed \emph{GM-SRM} first pre-learns a generative memory from the whole training data to explicitly learn the distribution of different semantic patterns. 
Then the learned memory are leveraged to retrieve the matching semantics for the current corrupted image to perform semantic reasoning during image inpainting. 
While the encoder-decoder network is used for guaranteeing the pixel-level content consistency, our generative priors are favorable for performing high-level semantic reasoning, which is particularly effective for inferring semantic content for large corrupted area.
Extensive experiments on Paris Street View, CelebA-HQ, and Places2 benchmarks demonstrate that our \emph{GM-SRM} outperforms the state-of-the-art methods for image inpainting in terms of both visual quality and quantitative metrics.

%% file: 1.introduction.tex
Image inpainting aims to infer the content of missing regions given a corrupted image. It serves as the essential technical step for many image processing tasks, such as old-photo restoration~\cite{wan2020bringing} and image edit~\cite{bau2019semantic}. Image inpainting is challenging in that the predicted content for the missing regions is required to be consistent with the known regions at both the pixel level and the semantic level. Despite the significant progress for image inpainting made in recent years, image inpainting for images with large corrupted area remains an extremely challenging task.

Compared to the traditional techniques~\cite{ballester2001filling,barnes2009patchmatch} for image inpainting, deep learning methods based on convolutional neural networks have boosted the performance of image inpainting substantially due to its excellent capability of feature learning. Most existing deep learning methods~\cite{ren2015shepard,pathak2016context} for image inpainting follow the encoder-decoder framework, in which an encoder is designed to extract useful features from the known regions of the input image, and then the decoder infers the content of the corrupted regions based on the encoded features. While such straightforward modeling way performs well for image inpainting with small corrupted regions, it can hardly deal with images with large corrupted area. This is largely because such methods focus on modeling the image-to-image mapping between the input corrupted image and the groundtruth intact image during training, whereas the reasoning from the known regions to the corrupted regions is not explicitly learned. 

To fully take advantage of the information of known regions for inferring the content of the corrupted regions, many methods for image inpainting~\cite{liu2018image,yu2019free,Li_2020_CVPR,Nazeri_2019_ICCV,ren2019structureflow,jie2020inpainting,Li_2019_ICCV,song2018contextual,Yan_2018_Shift,yu2018generative,wang2019musical,liu2019coherent,xie2019image,wang2020multistage,zheng2019pluralistic,zhao2020uctgan} aim to learn effective prior knowledge from the known regions of the input image, and then infer the content of the corrupted regions based on such prior knowledge. These methods can be classified into four categories by the way of leveraging priors from the known regions. The first type of methods~\cite{liu2018image,yu2019free,Li_2020_CVPR} employs attention mechanism to learn a confidence mask for the corrupted regions based on prior information of known regions, and then infers the pixel values of corrupted regions in a progressive propagation manner from high-confidence area to low-confidence area. The second way of leveraging priors of known regions~\cite{Nazeri_2019_ICCV,ren2019structureflow,jie2020inpainting,Li_2019_ICCV} is to predict the structure information of the corrupted regions from known regions first, namely the edge (high-frequency) information, then infer the detailed texture information under the guidance of the structure information. The prior knowledge of known regions can also be learned using VAE~\cite{kingma2013auto} by estimating the distribution of pixel values of the whole image based on known regions~\cite{zheng2019pluralistic,zhao2020uctgan}. Then the values of the corrupted regions can be inferred based on the obtained distribution. The last type of methods~\cite{song2018contextual,Yan_2018_Shift,yu2018generative,wang2019musical,liu2019coherent,xie2019image,wang2020multistage} performs constraints on the semantic consistency between the known regions and the predicted corrupted regions. 

\begin{figure*}[t]
\centering
\includegraphics[width=1.0\linewidth]{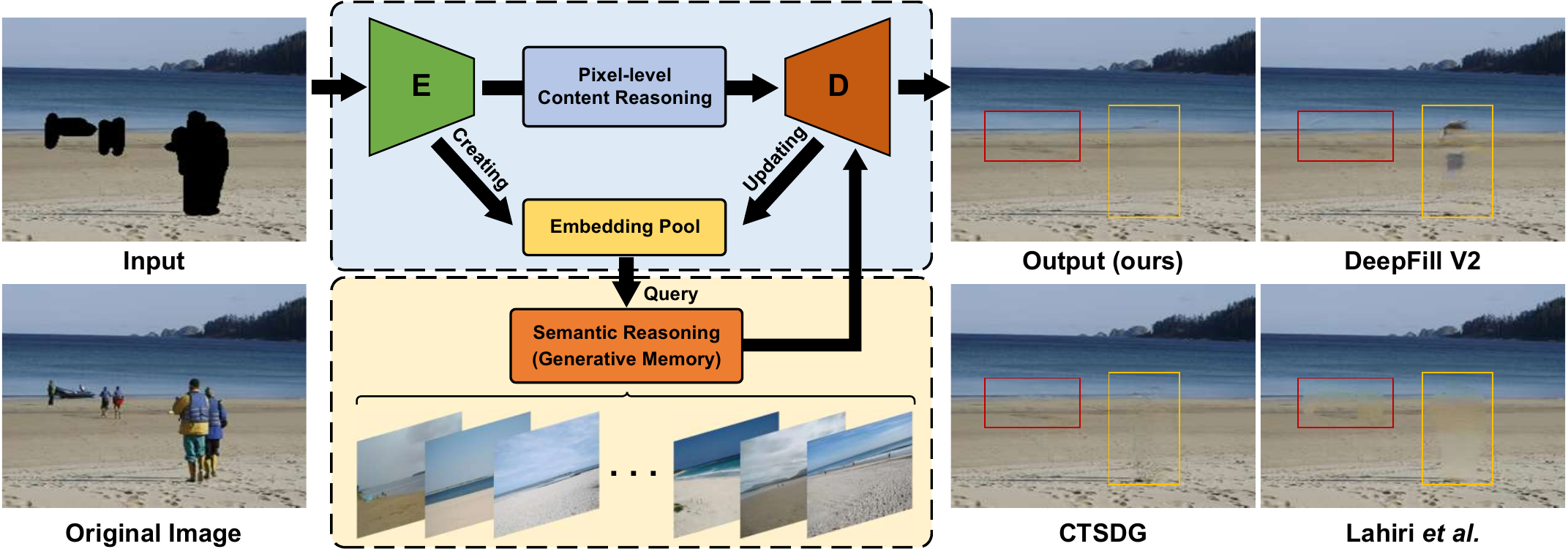}
\vspace{-4pt}
\caption{Given a corrupted image, our \emph{GM-SRM} infers the content for the corrupted area by leveraging two types of content reasoning by the decoder (\textbf{D}): 1) pixel-level content reasoning by the encoder-decoder framework, and 2) high-level semantic reasoning by the pre-trained generative memory.
Our \emph{GM-SRM} obtains the semantic priors from the whole training data by learning a generative memory, and the learned priors are queried by the embeddings of known information for high-level semantic reasoning. 
Consequently, our model is able to inpaint more reasonable content than other state-of-the-art methods, such as DeepFill V2~\cite{yu2019free} which implicitly learns the inter-image semantic priors, CTSDG~\cite{Guo_2021_ICCV} which uses a two-stream network to facilitate image inpainting by the structure-constrained texture synthesis and texture-guided structure reconstruction, or Lahiri~\emph{et al.}~\cite{lahiri2020prior} which explicitly learns inter-image priors by simply pre-training the decoder as an image synthesizing network.}
\vspace{-4pt}
\label{fig:pip}
\end{figure*}

All aforementioned methods focus on modeling the image-to-image mapping and implicitly learn semantic priors from the known regions of current input image to infer the content of the corrupted regions. 
Such modeling paradigm suffers from two potential limitations:
1) Such one-to-one learning paradigm tends to overfit each single training pair of images;
2) The inter-image prior knowledge about the general distribution patterns of visual semantics, which can be transferred across images sharing similar semantics, is not explicitly exploited. 
For instance, similar objects or scenes from different but similar images always share similar textures. 
Besides, the spatial/structural associations among objects/sub-regions can also be distilled as priors to generalize across similar semantics.
Lahiri \emph{et al.}~\cite{lahiri2020prior} made the first heuristic attempt to explicitly learn such semantic prior by pre-training a plain Generative Adversarial Net (GAN)~\cite{NIPS2014_5ca3e9b1} as the decoder to reconstruct the corrupted image.
The pre-trained GAN captures the general semantic distribution, which is used to predict the content of the corrupted regions during decoding. 
However, such a straightforward way suffers from three main weaknesses for handling large corruptions.
1) The pre-trained decoder is hard to synthesize exquisite details only with semantic reasoning;
2) The manner of semantic query solely relies on one highly coupled embedding to control all types of semantics for different feature scales; 
3) Vanilla GAN has limited generative performance to provide complex semantics for largely corrupted regions.
As a result, this method only shows effectiveness on images with small corrupted regions, but cannot handle cases with large corrupted regions.

In this paper, we propose the \textbf{G}enerative \textbf{M}emory-guided \textbf{S}emantic \textbf{R}easoning \textbf{M}odel (\emph{GM-SRM}), which infers the content of corrupted regions based on not only the known regions of the corrupted image, but also the learned inter-image reasoning priors characterizing the generalizable semantic distribution patterns between similar images.
Similar to most existing methods, the pixel-level reasoning from the known regions of the input image learns the image-to-image mapping following the typical encoder-decoder framework.
To obtain more comprehensive semantic priors, our \emph{GM-SRM} mines the general distribution of semantic patterns
that can be transferred across images sharing similar semantic distributions. 
Specifically, our proposed \emph{GM-SRM} first pre-learns a generative memory from the whole corpus of training data to capture the semantic distributions in a global view. Then the learned memory is leveraged to guide the process of image inpainting: it is favorable for performing high-level semantic reasoning while the typical encoder-decoder framework focuses on guaranteeing mainly the low-level (like pixel-level) semantic consistency.
As shown in Figure~\ref{fig:pip}, our model is able to synthesize more reasonable content for the corrupted regions than competing methods. The main contributions of our \emph{GM-SRM} are summarized as follows:
\begin{itemize}
    \item A novel image inpainting framework is proposed to learn both pixel-level content reasoning and semantic reasoning by the generative prior knowledge.
    It employs the inter-image priors from other images sharing similar semantic distributions to facilitate inpainting reasonable content in the corrupted area. 
    \item We design a GAN-based generative memory to learn the generative prior knowledge about the general distribution of semantic patterns,
    which can be seamlessly integrated into the typical encoder-decoder framework.
    \item We present the Conditional Stochastic Variation mechanism to learn the texture distribution
    of the known regions, thereby synthesizing rich yet semantically reasonable textures during image inpainting.
    \item Extensive experiments on three benchmarks of image inpainting, including Paris Street View, CelebA-HQ, and Places2, demonstrate both the quantitative and qualitative superiority of our proposed \emph{GM-SRM} over other state-of-the-art methods for image inpainting, especially in scenarios with large corrupted area.
\end{itemize}

%% file: 2.Relatedwork.tex

There is a substantial amount of work on image inpainting. In this section, we review the most typical methods. Mainstream image inpainting methods can be roughly divided into two categories: non-learning based methods and learning-based methods.

\subsection{Non-learning based Image Inpainting}
Earlier research on image inpainting employs the content of existing regions to directly fill in the missing regions, which can be divided into two categories, diffusion-based methods, and patch-based methods. 
The diffusion-based methods~\cite{ballester2001filling,bertalmio2000image} extend contextual pixels from the boundary to the hole along the isophote direction.
By imposing various boundary conditions, diffusion-based methods have shown promising performance in filling reasonable content in images with small corruptions.
However, diffusion-based inpainting methods often cause boundary-related blurriness and synthesize implausible semantics.

Aiming to solve such problems, the patch-based methods~\cite{barnes2009patchmatch,efros2001image,kwatra2005texture,criminisi2004region,jin2018patch} copy and paste the most similar patch from known regions or external images to replace the corrupted regions. Thus, many patch-based methods contribute to designing the optimal algorithms for patch selection.
For instance, Criminisi \emph{et al.}~\cite{criminisi2004region} propose an exemplar-based method for texture synthesis and calculate the confidence value for determining the inpainting order of the corrupted region.
Barnes \emph{et al.}~\cite{barnes2009patchmatch} propose the PatchMatch algorithm for quickly selecting approximate nearest neighbor matches between image patches.
Later, Jin~\emph{et al.}~\cite{jin2018patch} propose a patch-sparsity-based image inpainting algorithm through facet deduced directional derivative.
This type of inpainting methods can produce high-quality results on images with repeated textures. Nevertheless, it generates significant artifacts in the output due to the lack of global semantic understanding and generalization. In summary, non-learning based inpainting methods are difficult to satisfy the demand of practical application nowadays.

\subsection{Learning-based Image Inpainting}
With the great success of deep convolutional neural networks(CNNs) in various computer vision tasks, recent research leverages 
the encoder-decoder CNN framework to facilitate image inpainting. Main CNN-based methods for image inpainting employ known regions as priors to infer missing pixels from the outside to the inside.
Pathak \emph{et al.}~\cite{pathak2016context} firstly propose a context-encoder framework to restore corrupted images in latent space. 
To further cope with irregular corruption, Liu \emph{et al.} introduce the partial convolution(PConv)~\cite{liu2018image} to avoid blurry artifacts caused by traditional convolution. 
Inspired by attention mechanism~\cite{vaswani2017attention}, Yu \emph{et al.} propose gated convolution(DeepFill V2)~\cite{yu2019free} to gradually learn the soft mask rather than the hard mask in PConv, enhancing the inference reliability.
Yu \emph{et al.} believe features of known regions and inferred regions should be normalized respectively, thus proposing the region normalization(RN)~\cite{yu2020region} to improve inpainting performance.
Li \emph{et al.}~\cite{Li_2020_CVPR} propose a recurrent framework to iteratively predict the features of corrupted regions and merge them to infer the corrupted content.

\smallskip\noindent\textbf{Structural prior-guided inpainting methods.}
To synthesize more plausible results, some research employs structure information as prior knowledge to reconstruct unknown regions.
For instance, Liu \emph{et al.}~\cite{liu2019coherent} improve the attention mechanism by computing coherent semantic relevance to infer missing content. 
Nazeri~\emph{et al.}~\cite{Nazeri_2019_ICCV} predict edge maps to guarantee structure correctness and reuse edge maps to facilitate inpainting plausible content.
Yang~\emph{et al.}~\cite{jie2020inpainting} further integrate structural information with semantic information to enhance stability of image inpainting.
Xiong~\emph{et al.}~\cite{Xiong_2019_CVPR} propose a foreground-aware framework that restores foreground objects of input image.
Liu~\emph{et al.}~\cite{Liu2019MEDFE} propose a mutual encoder-decoder framework to equalize the restoration of structure features and texture features.
Xu~\emph{et al.}~\cite{xu2020e2i} design a two-step framework, which first generates edges inside the missing areas,
and then generates inpainted image based on the edges.
Wang~\emph{et al.}~\cite{wang2020structure} introduce a structure-guided method for video inpainting, which integrates scene structure, texture, and motion to complete the missing regions.
Recently, Guo~\emph{et al.}~\cite{Guo_2021_ICCV} propose a two-stream network which casts image inpainting into two collaborative subtasks, structure-constrained texture synthesis and texture-guided structure reconstruction.
Though structural prior-based methods have improved the performance to restore large corruption, the error of predicting structural prior often leads to synthesizing unreasonable semantics and reducing the quality of filled content.


\smallskip\noindent\textbf{Generative prior-guided inpainting methods.}
Learning the distribution of semantic patterns by generative models leads to promising performance on image inpainting.
Promoted by excellent generative representation ability of generative adversarial nets~\cite{NIPS2014_5ca3e9b1} and variation auto-encoders~\cite{kingma2013auto}, some research attempts to employ them to learn generative prior for improving the quality of synthesized images, and generate diverse content to fill in corrupted regions.
Li~\emph{et al.}~\cite{li2017generative} propose a generative adversarial model to learn distribution representation of natural images, and thus can synthesize reasonable content from random noise.
Zheng~\emph{et al.}~\cite{zheng2019pluralistic} introduce the VAE to enhance the diversity of generated content by posterior probability maximization.
Later, Zhao~\emph{et al.}~\cite{zhao2020uctgan} propose a VAE-based framework to improve the diversity of filled content by cooperation with randomly selecting instance images.

\begin{figure*}[!t]
\centering
\includegraphics[width=1.0\linewidth]{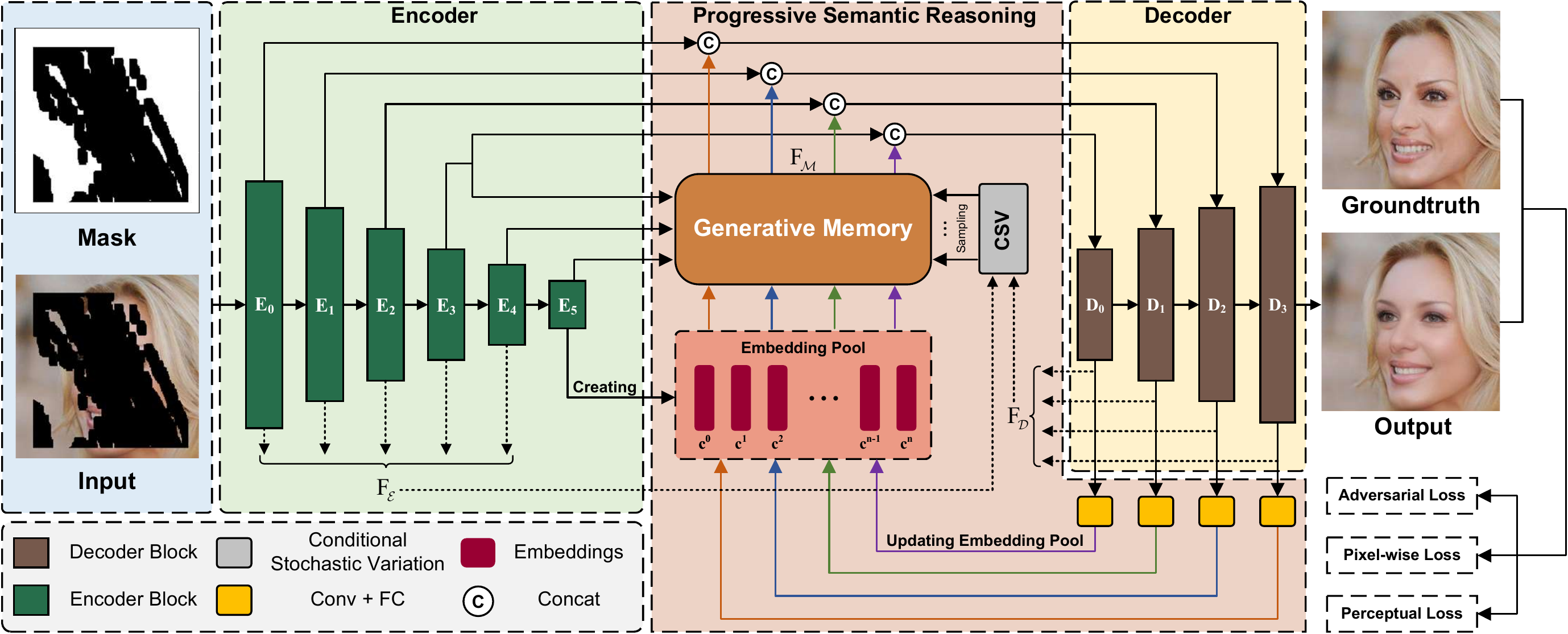}
\caption{The architecture of the proposed Generative Memory-guided Semantic Reasoning Model(\emph{GM-SRM}). 
It consists of three modules: the encoder $\mathcal{E}$, the generative memory $\mathcal{M}$, and the decoder $\mathcal{D}$. 
The proposed \emph{GM-SRM} infers the corrupted regions by two types of content reasoning: 1) the generative priors on the distribution of various semantic patterns, which are learned by the proposed generative memory $\mathcal{M}$ from the whole training corpus, and 2) pixel-level content reasoning by the encoder-decoder framework assisted by the semantics from the generative memory.
For each input corrupted image, the semantic priors are retrieved from the pre-trained generative memory $\mathcal{M}$ by searching for semantically matched features for the corrupted area using the encoded embedding $c$, representing known information.
The semantic priors by generative memory favor the high-level semantic reasoning for large corruption.
}
\label{fig:arch}
\end{figure*}
To explicitly learn the generative prior, Lahiri~\emph{et al.}~\cite{lahiri2020prior} pre-train a vanilla gan model as generative prior of the whole dataset, and then search effective information from the distribution space. 
Kelvin~\emph{et al.}~\cite{chan2020glean} propose the GLEAN framework to learn generative priors for synthesizing realistic textures in the image super-resolution task.
However, current generative prior-guided methods fail to restore images with complex content, or large corruption due to the weakness of low-level content inference. 
Hence, our \emph{GM-SRM} combines the pixel-level content reasoning in the typical encoder-decoder framework and high-level semantic reasoning by learning the generative prior from the proposed generative memory respectively.

%% file: 3.Method.tex
Given a corrupted image, the goal of image inpainting is to infer the content of the missing (corrupted) regions, which is required to be consistent with the known regions at both the pixel level and the semantic level. To this end, we propose the \textbf{G}enerative \textbf{M}emory-guided \textbf{S}emantic \textbf{R}easoning \textbf{M}odel (\emph{GM-SRM}), which learns a generative memory from the corpus of training data to mine the general distribution patterns of visual semantics within images. Such learned generative memory is then leveraged to guide the process of image inpainting by performing 
semantic reasoning for the missing regions.

In this section, we will first introduce the overall framework of image inpainting by our model, and then we will elaborate on how to construct the generative memory and perform semantic reasoning to infer the content of the missing regions. Next, we present the conditional stochastic variation, a technique proposed to synthesize rich yet semantically reasonable details during image inpainting.
Finally, we will show how to perform supervised learning to train our proposed \emph{GM-SRM}.

\subsection{Overall Framework for Image Inpainting}
As illustrated in Figure~\ref{fig:arch}, our \emph{GM-SRM} mainly consists of three modules: the encoder $\mathcal{E}$, the generative memory $\mathcal{M}$, and the decoder $\mathcal{D}$. The whole model follows the classical encoder-decoder framework to perform image inpainting. The encoder $\mathcal{E}$ is responsible for extracting useful visual features for the known regions of the input corrupted image and meanwhile generating a latent embedding which serves as a query embedding to search for semantically matched visual features from the generative memory $\mathcal{M}$. The generative memory $\mathcal{M}$ is constructed to learn the general distribution patterns of visual semantics from training data and help infer the content of the missing regions that is semantically consistent with the known regions in the input image. Thus the generative memory $\mathcal{M}$ summarizes the prior knowledge about the distribution of high-level visual semantics within images from a global view (covering the whole corpus of training data). Such a generative prior knowledge enables our model to explicitly infer more 
semantically reasonable content for the missing regions than the typical way, which trains the deep model to implicitly learn the consistency between similar semantics.
The retrieved semantically matched features from the memory $\mathcal{M}$, together with the encoded features from the encoder $\mathcal{E}$, are fed into the decoder $\mathcal{D}$ to synthesize the intact image. 

\smallskip\noindent\textbf{Encoder.} Given a corrupted image $I$ with a mask $M$ (with the same size as $I$) indicating the missing regions, the encoder $\mathcal{E}$ extracts visual features for the known regions of $I$ by:
\begin{equation}
    \mathbf{F}_{\mathcal{E}} = \mathcal{E}(I, M),
\end{equation}
where $\mathbf{F}_{\mathcal{E}}$ denotes the obtained encoded features.
As shown in Figure~\ref{fig:arch}, the encoder $\mathcal{E}$ is designed by repetitively stacking a basic residual block to iteratively refine visual features and meanwhile downsample the resolution of feature maps. Such residual block consists of stride-2 convolution layers along with ReLU layers. Besides, we utilize instance normalization (IN)~\cite{ulyanov2016instance} and channel attention (CA) mechanism~\cite{hu2018squeeze} to optimize the feature learning process. Specifically, the $i$-th basic residual block in the encoder $\mathcal{E}$ can be mathematically formulated as: 
\begin{equation}
\begin{split}
    &\mathbf{F}^e = \text{ReLU}(\text{IN}(\text{Conv}(\mathbf{F}_{\mathcal{E}}^{i-1}))), \\
    &\mathbf{F}_{\mathcal{E}}^{i} = \mathbf{F}^e + \text{CA}(\text{ReLU}(\text{IN}(\text{Conv}(\mathbf{F}^e)))),
\end{split}
\end{equation}
where $\mathbf{F}_{\mathcal{E}}^{i}$ is the output feature maps by the $i$-th basic residual block of $\mathcal{E}$, and $\mathbf{F}^e$ is the intermediate features in the residual connection.

Besides learning visual features for the known regions of the input corrupted image, the encoder $\mathcal{E}$ also generates the latent embeddings to establish an updating embedding pool based on the encoded features, and the query embeddings are the representation of progressively renovated known information. The embedding pool performs feature query by embeddings on the generative memory $\mathcal{M}$ to search for semantically matched features for the missing regions:
\begin{equation}
    \mathbf{c} = f_{c} (\mathbf{F}_{\mathcal{E}}),
    \label{eqn:latent_code}
\end{equation}
where the latent embedding $\mathbf{c}$ is a vector obtained by a non-linear mapping function $f_{c}$ performed by a convolutional layer and a fully connected layer from the encoded features $\mathbf{F}_{\mathcal{E}}$.

\begin{figure}[!t]
\centering
\includegraphics[width=0.99\linewidth]{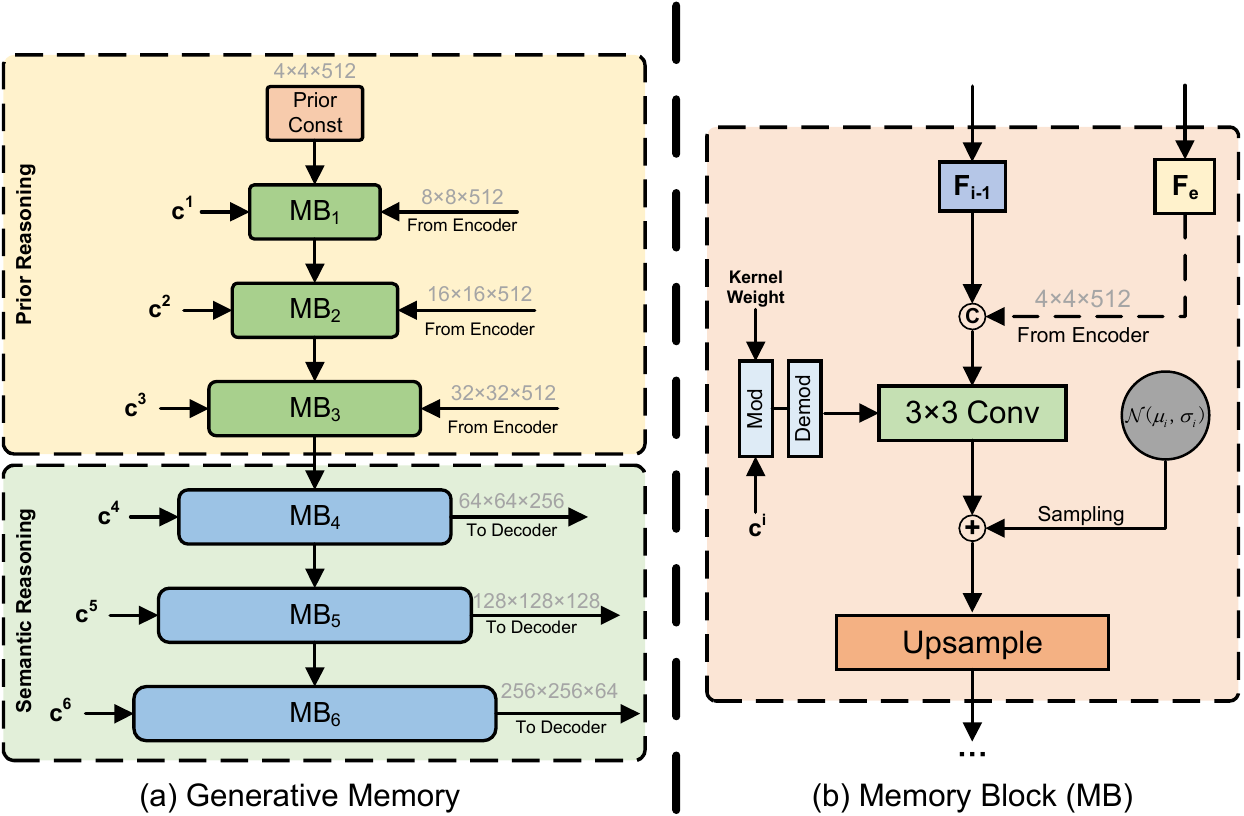}
\caption{The structure of the proposed generative memory in \emph{GM-SRM}. $c^i$ denotes the latent embedding, $F_e$ and $F_{i-1}$ represent the features from the encoder and last memory block respectively, and \textcircled{c} is the feature concatenation operation.}
\label{fig:memory}
\vspace{-4pt}
\end{figure}

\smallskip\noindent\textbf{Generative Memory.} The generative memory $\mathcal{M}$ is designed as a distribution learner by a StyleGAN-based generative network to infer semantics in corrupted regions from the latent vectorial embedding distilled by the features of known regions.
The main goal of the generative memory $\mathcal{M}$ is to synthesize the matched feature maps that are semantically consistent with the known regions, under the supervised learning for the whole \emph{GM-SRM} model.
It takes the updating latent embedding $\mathbf{c}$ as well as the encoded features $\mathbf{F}_{\mathcal{E}}$ as input and generates the corresponding feature maps. 
As a result, the generative memory $\mathcal{M}$ learns a mapping from the latent coding space to the reasoned semantic features. 
Formally, the reasoned semantic features $\mathbf{F}_{\mathcal{M}}$ corresponding to the latent embedding $\mathbf{c}$ by the generative memory $\mathcal{M}$ is queried by:
\begin{equation}
    \mathbf{F}_{\mathcal{M}} = \mathcal{M}(\mathbf{c}, \mathbf{F}_{\mathcal{E}}).
\end{equation}
We will elaborate on the model structure of the generative memory $\mathcal{M}$ and describe how to perform progressive semantic reasoning based upon $\mathcal{M}$ in the following Section~\ref{sec:memory}.

\begin{algorithm}[t]
\setstretch{1.1}
\begin{small}
    \caption{Inference Process of \emph{GM-SRM}}
    \label{alg:gm-srm}
    \hspace*{0.015in}{\bf Input:}
    $\text{I}_{in},\text{M}_{in},\mathcal{M},\text{E}_i,\text{D}_i,\text{c}.$\\
    $\text{I}_{in}:$ Input corrupted image;\\
    $\text{M}_{in}:$ Mask of corrupted region;\\
    $\mathcal{M}:$ Pre-trained generative memory;\\
    $\text{E}_i:$ i-th layer of encoder $E$;\\
    $\text{D}_i:$ i-th layer of decoder $D$;\\
    $\text{c}:$ latent embeddings for memory reasoning.\\
    \hspace*{0.015in}{\bf Output:}
    $\hat{\text{I}}_{out}$, restored image from input $\text{I}_{in}$.
    \begin{algorithmic}[1]
        \State \# Initializing input; \\
        $\text{F}^1_{\mathcal{E}}$ = Concat[$\text{I}_{in},\text{M}_{in}$] 
        \State \# Encoding, $n_e$ is the layer number of encoder; 
        \For{i = 1 $\rightarrow$ $n_e-1$}
            \State$\text{F}^{i+1}_{\mathcal{E}}$ = $\text{E}_i(\text{F}^i_{\mathcal{E}})$ 
            \State\# Estimating the mean $\mu$ and the standard variation $\sigma^2$; 
            \State $\mu_i$, $\sigma^2_i$ = \text{Mean(Split}($\text{Conv}$($\text{F}^i_{\mathcal{E}}$))) 
        \EndFor
        \State \textbf{end for}
        \State \# Initializing query embedding pool;
        \State $\text{c}_1$ = $\text{Linear}$($\text{Conv}$($\text{F}^{n_e}_{\mathcal{E}}$))
        \State $\text{F}^1_{\mathcal{D}}=\text{F}^{n_e}_{\mathcal{D}}$
        \State \# Decoding, $n_d$ is the layer number of decoder;
        \For{j = 1 $\rightarrow$ $n_d-1$}
            \State${\text{F}^{j+1}_{\mathcal{D}}}$ = $\text{D}_j$($\text{F}^j_{\mathcal{D}}$)
            \State \# Updating the latent embedding pool;
            \State$c_{j+1}$ = $\text{Linear}$($\text{Conv}$($\text{F}^{n_d-j}_{\mathcal{D}}$)) + $c_j$ 
            \State Sampling noise$_{j+1}$ from distribution $\mathcal{N}$($\mu_{n_d-j}$, $\sigma^2_{n_d-j}$)
            \State \# Semantic reasoning in generative memory;
            \State $\text{F}_{\mathcal{M}}$ = $\mathcal{M}$($\text{F}^{n_d-j}_{\mathcal{E}}$, $\text{c}_{j+1}$, \text{noise}$_{j+1}$)
            \State \# Inferring by known content and memory query;
            \State ${\text{F}^{j+1}_{\mathcal{D}}}$ = Concat[$\text{F}^{j+1}_{\mathcal{D}}$, $\text{F}_{\mathcal{M}}$]
        \EndFor 
        \State \textbf{end for}
        \State\# Synthesizing final restored output image.
        \State$\hat{\text{I}}_{out}$ = $\text{OutConv}({\text{F}^{n_d}_{\mathcal{D}}}$) 
    \end{algorithmic}
\end{small}
\end{algorithm}

\smallskip\noindent\textbf{Decoder.} The queried features $\mathbf{F}_{\mathcal{M}}$ can be considered as the inferred semantic features as generative priors by the memory $\mathcal{M}$ for the missing regions of the input corrupted image. 
Both $\mathbf{F}_{\mathcal{M}}$ and the encoded features $\mathbf{F}_{\mathcal{E}}$ for the known regions as the pixel-level inference cues
are fed into the decoder $\mathcal{D}$ to synthesize the output intact image $\hat{I}$:
\begin{equation}
    \hat{I} = \mathcal{D}(\mathbf{F}_{\mathcal{M}}, \mathbf{F}_{\mathcal{E}}).
\end{equation}
The decoder $\mathcal{D}$ is built in the similar way as the encoder: repetitively stacking a basic residual block to progressively synthesize the final intact image. One major difference from the encoder is that the decoder utilizes the bi-linear interpolation $f_{\text{interp}}$ in each basic block to upsample feature maps gradually. Specifically, the $i$-th basic block of the decoder $\mathcal{D}$ is formulated as:

\vspace{-4pt}
\begin{small}
\begin{equation}
\begin{split}
    &\mathbf{F}^d = \text{ReLU}(\text{IN}(\text{Conv}(f_{\text{interp}}(\text{Concat}[\mathbf{F}_{\mathcal{M}}^i, \mathbf{F}_{\mathcal{E}}^i, \mathbf{F}^{i-1}_{\mathcal{D}}])))),\\
    &\mathbf{F}^i_{\mathcal{D}} = \mathbf{F}^d + \text{CA}(\text{ReLU}(\text{IN}(\text{Conv}((\mathbf{F}^d)))),
\end{split}
\label{eqn:decoder}
\end{equation}
\end{small}\noindent 
where $\mathbf{F}^i_{\mathcal{D}}$ is the output feature maps of the $i$-th basic block in the decoder $\mathcal{D}$, and $\mathbf{F}^d$ is the intermediate features in the residual connection.
The whole inference process of our \emph{GM-SRM} is summarized in Algorithm~\ref{alg:gm-srm}.

\subsection{Semantic Reasoning by Generative Memory}
\label{sec:memory}

The generative memory $\mathcal{M}$ is proposed to employ adversarial learning to explicitly learn the distribution of various semantic patterns, and facilitate semantic reasoning from limited information.
Our GM-SRM then leverages the learned prior knowledge about the semantic distributions by the generative memory to perform high-level semantic reasoning for the missing regions given the information of known regions.

\smallskip\noindent\textbf{Construction of Generative Memory.}
As shown in Figure~\ref{fig:memory}, we opt for the generative model of revised StyleGAN~\cite{karras2020analyzing} due to its excellent performance of synthesizing target image controlled by decoupled latent embeddings via the improved AdaIN operation with the proposed demodulation~\cite{karras2020analyzing}.
Inspired by previous effective attempts to learn generative priors in other tasks~\cite{chan2020glean}, we pre-train the generative memory on the whole training data in the similar training way as plain StyleGAN, except that during training GM-SRM we replace the mapping network in the StyleGAN with the mapping function in Equation~\ref{eqn:latent_code} to learn the underlying correspondence between the latent embedding (representing the semantics of the existing regions) and the reasoned semantic features for the missing regions. 
After sufficient training on the whole corpus of training data, the generative memory $\mathcal{M}$ is expected to learn a good mapping between the latent space and visual semantic features, thereby summarizing the prior knowledge about the distribution of various semantic patterns from a global view. 
Once the memory $\mathcal{M}$ is pre-trained, its parameters $\mathcal{M}$ are frozen, and $\mathcal{M}$ is used to perform semantic reasoning as generative prior knowledge during the training of the whole \emph{GM-SRM}, i.e., to infer the semantic features for the missing regions of the input image based on the latent embedding of the known regions.

\smallskip\noindent\textbf{Progressive Semantic Reasoning.}
We leverage the pre-trained generative memory $\mathcal{M}$ to perform progressive semantic reasoning and thereby assist the decoder $\mathcal{D}$ to conduct image inpainting. As shown in Equation~\ref{eqn:decoder}, the decoder $\mathcal{D}$ employs multiple basic blocks to synthesize the intact image in a coarse-to-fine manner by progressively expanding the resolution of the generated feature maps. Accordingly, the generative memory infers the semantics for the missing regions for each resolution of feature maps to keep pace with the synthesizing process of the decoder $\mathcal{D}$. As a result, the memory $\mathcal{M}$ and the decoder $\mathcal{D}$ are able to perform image inpainting collaboratively in an iterative way: the inferred semantic features by $\mathcal{M}$ are fed into $\mathcal{D}$ to provide high-level semantic cues for decoding the same resolution of feature maps, while the decoded feature maps are in turn provided for $\mathcal{M}$ to update the embedding pool for the next scale of semantic inferring in larger resolution:
\begin{equation}
   \mathbf{c}^i = f_{c}(\mathbf{F}_{\mathcal{D}}^{i-1}) + \mathbf{c}^{i-1},
\end{equation}
where $f_c$ corresponds to the same mapping function as in Equation~\ref{eqn:latent_code}. $\mathbf{c}^i$ denotes the latent embedding to retrieve the semantic features $\mathbf{F}_{\mathcal{M}}^i$ from $\mathcal{M}$, which is prepared for decoding $\mathbf{F}_{\mathcal{D}}^{i}$ by the $i$-th block of $\mathcal{D}$. As a result, the newly inferred features for the missing regions can be incorporated into the updated latent embedding to predict the semantic features for the next basic block of $\mathcal{D}$. The image inpainting is performed in such a progressively inferring mechanism to predict the content of missing regions by the generative memory $\mathcal{M}$ and the decoder $\mathcal{D}$ collaboratively.

Compared to the typical methods for image inpainting that implicitly learns the distribution of semantics by modeling the image-to-image mapping, the key benefit of our \emph{GM-SRM} is that the generative memory explicitly learns the semantic distributions as prior knowledge in a global view from the whole corpus of training data, which can be generalized across different images sharing similar semantic distributions. Such learned prior knowledge about the general semantic distributions enables our model to perform high-level semantic reasoning to infer more 
semantically reasonable content for the missing regions. As a result, our method is particularly effective in scenarios with large corrupted area.
\vspace{-8pt}

\subsection{Conditional Stochastic Variation}
\begin{figure}[t]
\centering
\includegraphics[width=0.95\linewidth]{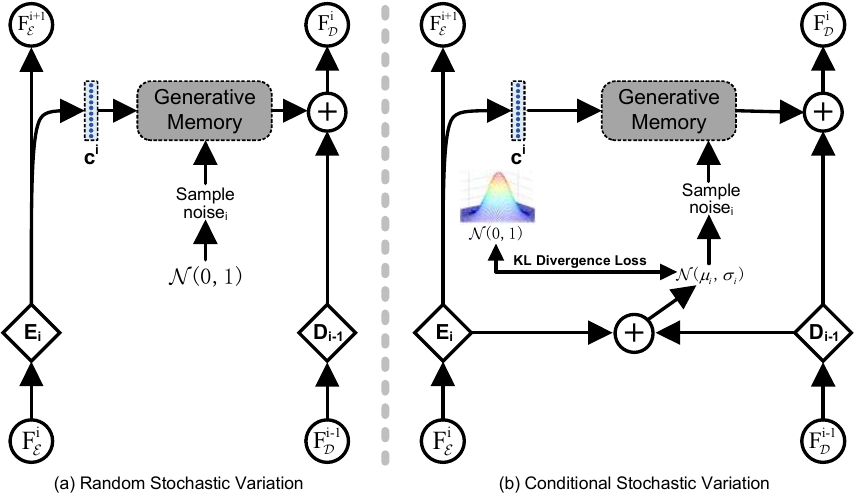}
\centering
\caption{Unlike typical StyleGAN that simply samples noise randomly from a standard normal distribution (a), our proposed Conditional Stochastic Variation mechanism samples noise conditioned on the encoded features $\text{F}_{\mathcal{E}}^i$ and the decoded features $\text{F}_{\mathcal{D}}^{i-1}$.}
\vspace{-8pt}
\label{fig:hs}
\end{figure}

To simulate the stochastic appearance variations in synthesized images that do not violate the correct semantics and enrich the texture details, random noise is introduced into synthesizing process in StyleGAN in multiple generative layers. Inspired by such Stochastic Variation scheme in StyleGAN, we also incorporate noise as input during image synthesis in the generative memory $\mathcal{M}$. Unlike the StyleGAN that synthesizes images from random latent embedding without conditioning on any input information, the generated features by $\mathcal{M}$ of our \emph{GM-SRM} model are required to be semantically consistent with the known regions. Thus we propose the Conditional Stochastic Variation mechanism, which introduces noise conditioned on the semantics of the known regions into the synthesis process of the generative memory $\mathcal{M}$ to enrich the appearance details of synthesized images while keeping the semantics correct.  

\begin{figure}[t]
\centering
		\includegraphics[width=0.85\linewidth]{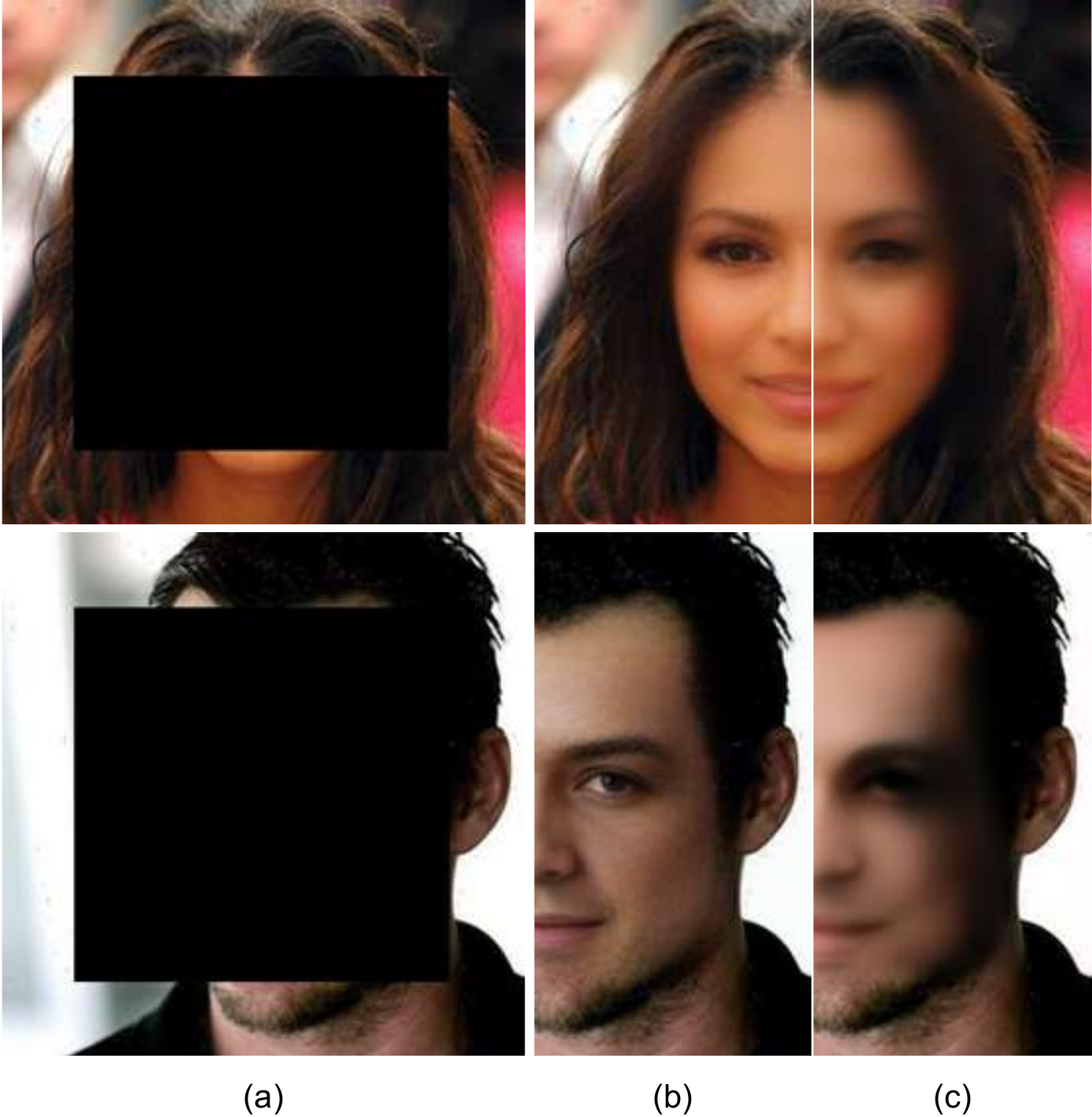}
\centering
\caption{Comparison of inpainting results between our \emph{GM-SRM} using random noise and our Conditional Stochastic Variation mechanism. (a) Corrupted input images. (b) Inpainted images using the conditional noise by the proposed Conditional Stochastic Variation mechanism. (c) Inpainted images using plain stochastic variation.}
\vspace{-4pt}
\label{fig:noise}
\end{figure}

Noise is introduced into each synthesizing layer of $\mathcal{M}$ to ensure the texture diversity of generated features in each resolution. As shown in Figure~\ref{fig:hs}, the noise in each layer is generated conditioned on the encoded features $\mathbf{F}_{\mathcal{E}}$ by $\mathcal{E}$ in the current layer and the decoded features $\mathbf{F}_{\mathcal{D}}$ by $\mathcal{D}$ in the previous layer. Specifically, a normal distribution is predicted from $\mathbf{F}_{\mathcal{E}}$ and $\mathbf{F}_{\mathcal{D}}$ by a convolution layer and a fully connected layer (FC), then the conditional noise is sampled from such predicted normal distribution. For instance,
the noise for the $i$-th layer of $\mathcal{M}$ is generated by the proposed Conditional Stochastic Variation Mechanism by:
\begin{equation}
\begin{split}
    &\mu_i, \sigma_i = \text{FC}(\text{Conv}(\mathbf{F}^i_{\mathcal{E}},\mathbf{F}^{i-1}_{\mathcal{D}}))), \\
    &\text{noise}_i = \text{Sample}(\mathcal{N}(\mu_i, \sigma_i^2)),
\end{split}
\label{eqn:normal_dist}
\end{equation}
where $\mu_i$ and $\sigma_i$ are the mean and standard variation for the predicted normal distribution for the $i$-th synthesis layer conditioned on the encoded features $\mathbf{F}^i_{\mathcal{E}}$ and the decoded features $\mathbf{F}^i_{\mathcal{D}}$. To encourage the generated noise distribution to be close to the standard normal distribution to ease the noise-sampling process, we leverage KL divergence to guide the parameter learning in Equation~\ref{eqn:normal_dist}, which is similar to VAE~\cite{kingma2013auto}:
\begin{equation}
    \mathcal{L}_{\text{KL}} = \sum w_i \cdot \text{KL}(\mathcal{N}(\mu_i, \sigma_i^2)~ |~ \mathcal{N}(0,1)),
    \label{eqn:KL}
\end{equation}
where $w$ for different layers are the weights to balance between different synthesis layers, and $\text{KL}$ denotes the KL divergence between two distributions.
Figure~\ref{fig:noise} shows two examples that illustrate the comparison between the typical Stochastic Variation mechanism performed by random noise and our proposed Conditional Stochastic Variation Mechanism. These examples reveal that the conditional noise generated by our Conditional Stochastic Variation leads to more consistent semantic content in the missing regions, while keeping rich texture details. In contrast, the random noise tends to result in blurred texture due to completely random nature of introduced noise.

\vspace{-4pt}
\subsection{Supervised Parameter Learning}
Our proposed \emph{GM-SRM} is trained in two steps: the generative memory $\mathcal{M}$ is first pre-trained to learn the prior knowledge about the general distribution patterns of visual semantics. Then the whole \emph{GM-SRM} is trained for image inpainting while keeping the parameters of $\mathcal{M}$ frozen. Since the generative memory $\mathcal{M}$ is trained in the similar way as the training process of StyleGAN, we explicate how to perform supervised learning to train the whole \emph{GM-SRM}.

We employ four types of loss functions to train our \emph{GM-SRM}. Apart from the loss function in Equation~\ref{eqn:KL}, the other three loss functions are presented below: 
\begin{itemize}[leftmargin =*]
\item \textbf{Pixel-wise L1 Reconstruction Loss},  which focuses on pixel-level measurement of known regions and corrupted regions between groundtruth and synthesized images respectively: 
\begin{equation}
\begin{split}
    \mathcal{L}_{\text{L1}} &= \|\hat{I} - I_{GT} \|_1, \\
    \mathcal{L}_{\text{rec}} &= \mathcal{L}_{\text{known-L1}} + \gamma \mathcal{L}_{\text{corrupted-L1}},
\end{split}
\end{equation}
where $\hat{I}$ is the synthesized image by our \emph{GM-SRM}, and $I_{GT}$ are the corresponding ground-truth image. $\gamma$ is a hyper-parameter to balance between losses, which is tuned to be 10 on a held-out validation set.

\item \textbf{Perceptual Loss}~\cite{johnson2016perceptual}, which aims to minimize the semantic difference between restored image and the ground-truth image in deep feature space:
\begin{equation}
    \mathcal{L}_{\text{perc}} (\hat{I}, I_{GT}) = \sum_{l=1}^L \frac{1}{C_l  H_l W_l} \| f^l_\text{vgg}(\hat{I}) - f^l_\text{vgg}(I_{GT})\|_1,
\end{equation}
where $f^l_\text{vgg}(\hat{I})$ and $f^l_\text{vgg}(I_{GT})$ are the extracted feature maps, normalized by feature dimensions $C_l \times H_l \times W_l$), for the generated image $\hat{I}$ and the ground-truth image $I_{GT}$ respectively from the $l$-th convolution layer of the pre-trained VGG-19 network~\cite{simonyan2014very}.

\noindent \item \textbf{Conditional Adversarial Loss}, which encourages the synthesized image $\hat{I}$ to be as realistic as the ground-truth image $I_{GT}$. We employ spectral normalization~\cite{miyato2018spectral} in discriminator to stabilize the training process:
\begin{equation}
    \mathcal{L}_{\text{adv}}  = -\mathbb{E}_{B\backsim \mathbb{P}_\text{GM-SRM}}[D^{sn}(G(I))], 
\end{equation}
where $D^{sn}$ is the spectral normalized discriminator. It is trained by:
\begin{equation}
\begin{split}
    \mathcal{L}_{D^{sn}}  = &\mathbb{E}_{B_{GT}\backsim \mathbb{P}_\text{data}}[1-D^{sn}(G(I))] \\ &+ \mathbb{E}_{B\backsim \mathbb{P}_\text{GM-SRM}}[D^{sn}(G(I))].
\end{split}
\end{equation}

\end{itemize}
Overall, the whole \emph{GM-SRM} is trained by:
\begin{equation}
    \mathcal{L} = \lambda_1\mathcal{L}_{\text{rec}} + \lambda_2\mathcal{L}_{\text{perc}}+\lambda_3\mathcal{L}_{\text{adv}}+\lambda_4\mathcal{L}_{\text{KL}},
\end{equation}
where $\lambda_1$, $\lambda_2$, $\lambda_3$ and $\lambda_4$ are hyper-parameters to balance between different losses. In our experiments, we empirically set $\lambda_1$=1, $\lambda_2$=0.1, $\lambda_3$=0.01 and $\lambda_4$=0.01.

%% file: 4.Experiments.tex
\input{Tables/table1}
In this section, we conduct experiments to quantitatively and qualitatively evaluate the proposed \emph{GM-SRM}. In Section~\ref{sec:metrics}, we introduce benchmark datasets used in the experiments, evaluation metrics, and implementation details. In Section~\ref{sec:comparison}, we conduct experiments to compare our model with state-of-the-art methods for image inpainting. Finally, we perform ablation study to investigate the effectiveness of each component in our \emph{GM-SRM} in Section~\ref{sec:ablation}. 
\subsection{Experimental Setting}
\smallskip\noindent\textbf{Evaluation Metrics.}
\label{sec:metrics}
We employ five generally used criteria as quantitative measurements to quantify the quality of the restored images in our experiments: 1) the peak signal-to-noise ratio (PSNR), 2) the structural similarity index (SSIM)~\cite{wang2004image}, 3) the normalized cross correlation (NCC)~\cite{wei2019single}, 4) the local mean square error (LMSE)~\cite{grosse2009ground}, and the learned perceptual image patch similarity (LPIPS)~\cite{zhang2018perceptual}. Additionally, we perform qualitative evaluation by visually comparing the restored results of randomly selected test samples by various models for different degrees of corruption in experiments. As a complement to the standard evaluation metrics, we further conduct user study to compare our results to the state-of-the-art results by human evaluation.

\smallskip\noindent\textbf{Datasets.} We perform experiments on three benchmark datasets for image inpainting:
\begin{itemize}[leftmargin=*]
\item \textbf{Paris Street View}~\cite{doersch2012makes}, which is collected from street views of Paris, and we leverage its original splits, 14,900 images for training and 100 images for testing.

\item \textbf{CelebA-HQ}~\cite{karras2018progressive}, which contains 30,000 images of human face. We randomly select 3000 images as validation and testing dataset, and leverage remained 27,000 images as training dataset.

\item \textbf{Places2}~\cite{zhou2017places}, which is composed of over 2,000,000 images from 365 scenes. We select two full categories obtaining 80,000 images and randomly picks 1,500 images from each category as test set respectively. The remaining 74,000 images are employed as training set.
\end{itemize}
By following previous experience~\cite{zheng2019pluralistic}, we randomly generate irregular and regular masks for training. As for test, we leverage PConv's~\cite{liu2018image} irregular mask and center mask in different degradation ratios.

\begin{figure*}[!t]
  \centering 
  \begin{minipage}[b]{0.975\linewidth}
  \includegraphics[width=\linewidth]{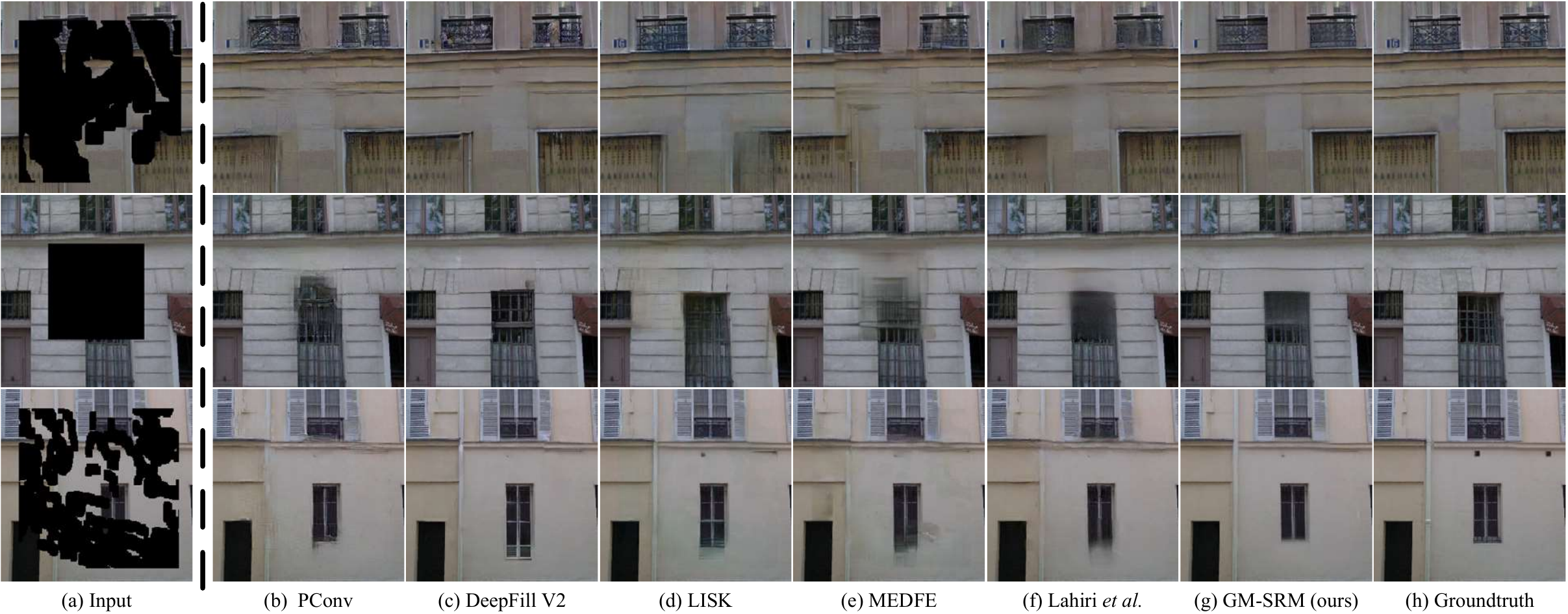}
  \end{minipage}
  \vspace{-8pt}
\caption{Visualization of inpainted images by five state-of-the-art models for image inpainting and our \emph{GM-SRM} on three randomly selected samples from Paris Street View~\cite{doersch2012makes} test set. Our model is able to restore higher-quality image than other methods. Best viewed in zoom-in mode.}
\vspace{-8pt}
\label{fig:case1}
\end{figure*}

\smallskip\noindent\textbf{Implementation Details.}
We implement our \emph{GM-SRM} in distribution mode with 4 RTX 3090 GPUs under Pytorch framework. Adam~\cite{kingma2014adam} is employed for gradient descent optimization with batchsize set to be 8. The initial learning rate is set to be $2\times 10^{-4}$ and the training process takes maximally 100 epochs. In our experiments, we resize all images to make the shorter side be 320, and then crop into 256$\times$256. Random flipping, random cropping, and resizing are used for data augmentation. 

\begin{figure}[t]
\centering
		\includegraphics[width=1.0\linewidth]{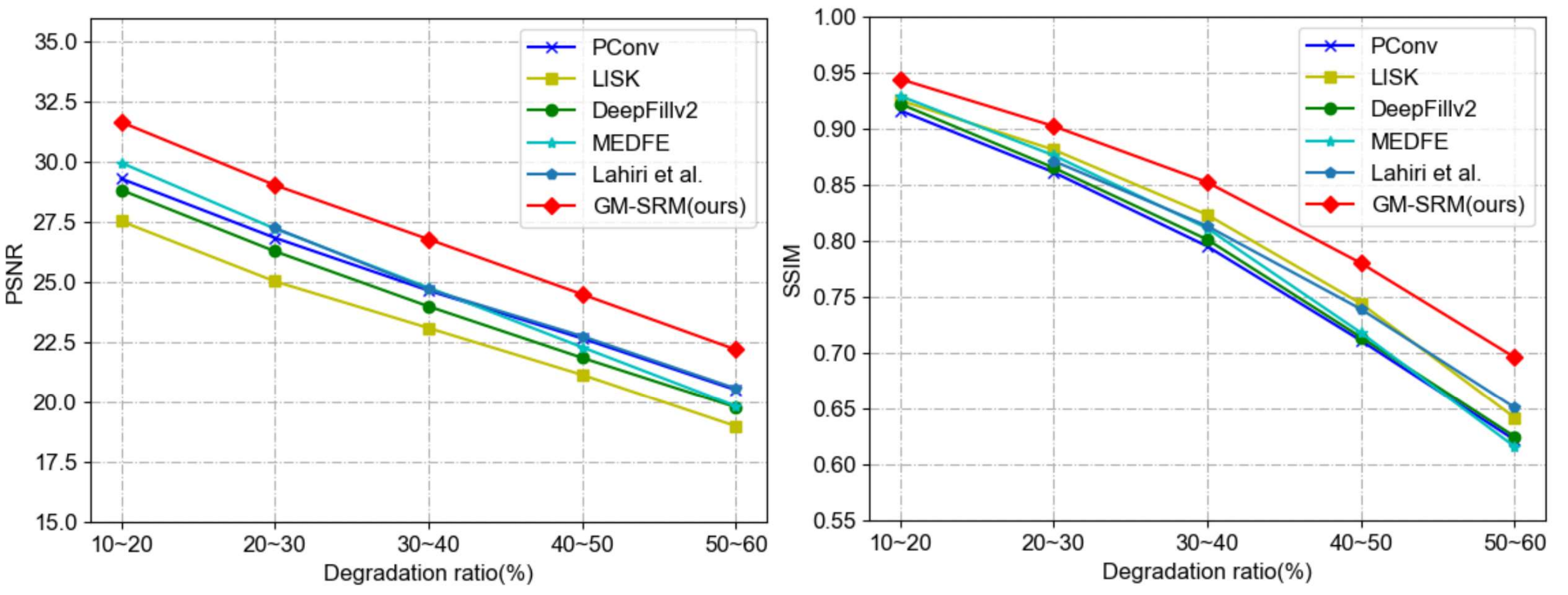}
\vspace{-20pt}
\centering
\caption{Quantitative comparison on Paris Street View~\cite{doersch2012makes} for varying degradation ratios in terms of PSNR and SSIM. Competing methods includes PConv~\cite{liu2018image}, DeepFillv2~\cite{yu2019free}, LISK~\cite{jie2020inpainting}, MEDFE~\cite{Liu2019MEDFE}, Lahiri \emph{et al.}~\cite{lahiri2020prior}, and our \emph{GM-SRM}. Though higher degradation ratio leads to faster decreasing of quantitative metrics, the \emph{GM-SRM} performs more stably than other state-of-the-art methods.} 
\vspace{-8pt}
\label{fig:fig1}
\end{figure}

\subsection{Comparison with State-of-the-art Methods}
\label{sec:comparison}
We first conduct experiments to compare \emph{GM-SRM} with state-of-the-art methods for image inpainting on three datasets, including Paris StreetView~\cite{doersch2012makes}, CelebA-HQ~\cite{karras2018progressive}, and Places2~\cite{zhou2017places}. In particular, we divide masks by ranging percentage of corrupted size. 

\smallskip\noindent\textbf{Baselines.} Concretely, we compare our \emph{GM-SRM} with 1) \textbf{PConv}~\cite{liu2018image}, employing partial convolution to cope with irregular corruption; 2) \textbf{DeepFill V2}~\cite{yu2019free}, which proposes gated convolution to generalize PConv; 3) \textbf{EdgeConnect}~\cite{Nazeri_2019_ICCV}, which first predicts edge map and then leverages predicted edge maps to facilitate restoration; 4) \textbf{LISK}~\cite{jie2020inpainting}, which incorporates structural knowledge to reconstruct corrupted image and structure maps simultaneously; 5) \textbf{LBAM}~\cite{xie2019image}, introducing a learnable reverse attention mechanism to fill missing regions;  6) E2I~\cite{xu2020e2i}, which designs a two-step framework to first generate edges inside the missing areas, and then generate inpainted image based on the edges;
7) \textbf{MEDFE}~\cite{Liu2019MEDFE}, which proposes a mutual encoder-decoder framework to reconstruct structure and textures separately, and then fuses them by feature equalization; 8) CTSDG~\cite{Guo_2021_ICCV}, which proposes a two-stream network which casts image inpainting into structure-constrained texture synthesis and texture-guided structure reconstruction; 9) \textbf{Lahiri \emph{et al.}}~\cite{lahiri2020prior}, which 
explicitly learns the generative priors by pre-training a vanilla GAN as the decoder.
It should be noted that all these methods except Lahiri \emph{et al.} implicitly learn the distribution of different semantics
to perform image inpainting.

\input{Tables/table2}

\smallskip\noindent\textbf{Quantitative Evaluation.} Table~\ref{tab:irregular} and Table~\ref{tab:center} present the experimental results of different methods for image inpainting on three benchmarks in terms of PSNR, SSIM, NCC, LMSE, and LPIPS. To quantify model performance, we leverage two types of corrupted masks for testing, regularly center masks and irregular masks. In addition, we divide them into various corrupted ratios, \emph{i.e.}, (20\%, 30\%], (30\%, 40\%], (40\%, 50\%], and (50\%, 60\%] for irregular mask, and 25\% and 50\% for center mask. For a fair comparison, we obtain restoration results from officially released source codes and pre-trained models. According to their source code and paper, we re-implement models that do not release official pre-trained models. We leverage the same mask for each test image to evaluate the results of different methods. 

    \begin{figure*}[!t]
      \centering 
  \begin{minipage}[b]{0.975\linewidth}
  \includegraphics[width=\linewidth]{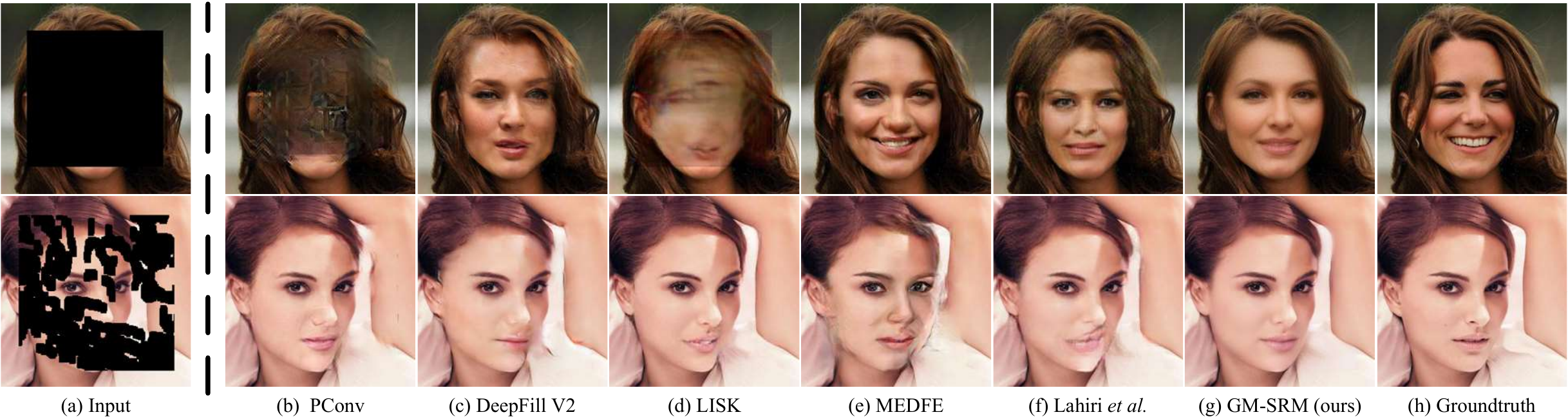}
  \end{minipage}
  \vspace{-4pt}
\caption{Visualization of inpainted images by five state-of-the-art models for image inpainting and our \emph{GM-SRM} on two randomly selected samples from CelebA-HQ~\cite{karras2018progressive} test set. Our model is able to restore higher-quality image than other methods. Best viewed in zoom-in mode.}
\vspace{8pt}
\label{fig:case2}

\centering 
  \begin{minipage}[b]{0.975\linewidth}
  \includegraphics[width=\linewidth]{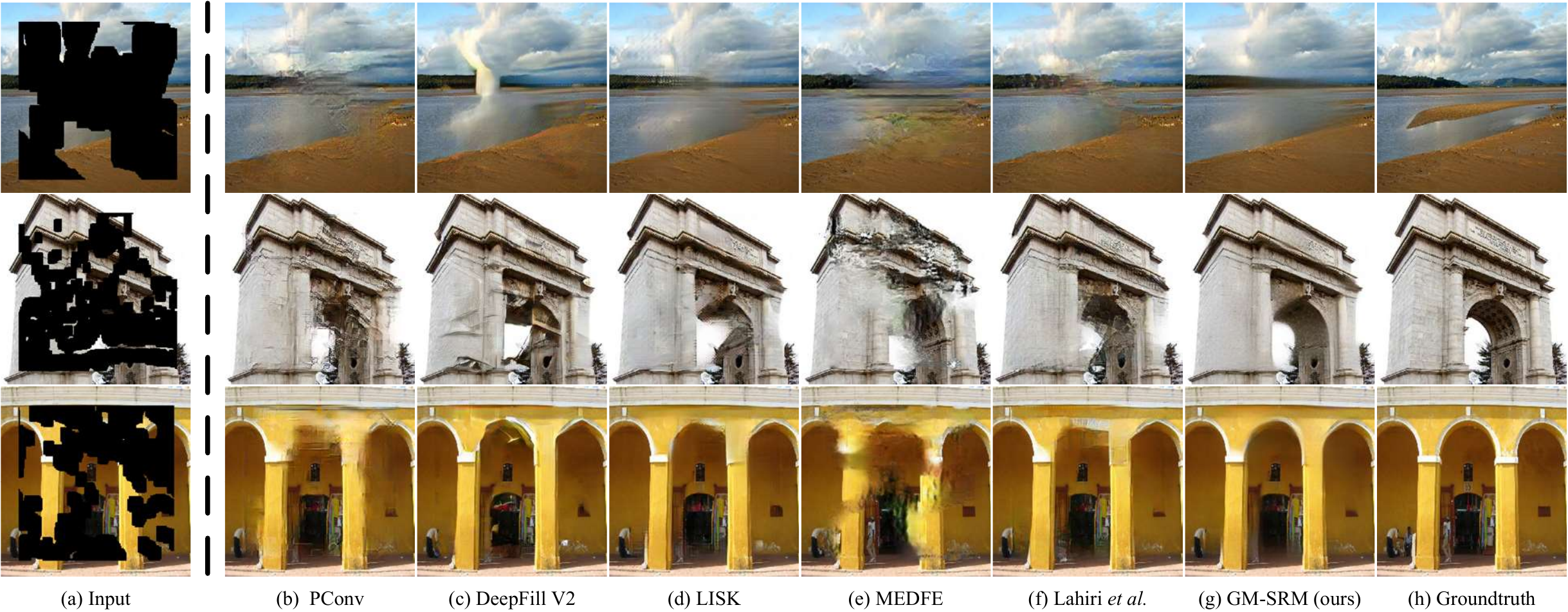}
  \end{minipage}
  \vspace{-4pt}
\caption{Visualization of inpainted images by five state-of-the-art models for image inpainting and our \emph{GM-SRM} on three randomly selected samples from Places2~\cite{zhou2017places} test set. Our model is able to restore higher-quality image than other methods. Best viewed in zoom-in mode.}
\vspace{-12pt}
\label{fig:case3}
\end{figure*}

As illustrated in Table~\ref{tab:irregular}, we compare quantitative results of our \emph{GM-SRM} with state-of-the-art methods on three benchmarks for irregular inpainting. On Paris Street View dataset~\cite{doersch2012makes}, our \emph{GM-SRM} outperforms other competing methods by a large margin on five metrics. As illustrated in Figure~\ref{fig:fig1}, our \emph{GM-SRM} achieves significant improvements (higher than 1.3 dB PSNR) for each corruption ratio. For CelebA-HQ dataset, although face structure is relatively fixed, our \emph{GM-SRM} still outperforms other state-of-the-art methods more than 1 dB of PSNR gain for the large hole-to-image area ratio (50\%,60\%]. The performance of our \emph{GM-SRM} on Places2~\cite{zhou2017places} dataset is also better than other methods, especially in the cases with large corruption ratios. 
In sum, our \emph{GM-SRM} achieves the best performance in terms of five quantitative metrics for irregular corruption, and outperforms other state-of-the-art methods in various corruption ratios. Comparing with the methods that implicitly learn semantic distribution, namely all methods in comparison except Lahiri \emph{et al.}, our method also explicitly learns the semantic priors by the proposed generative memory, which enables our model to perform high-level semantic reasoning between images with similar semantic distributions. It is worth noting that our \emph{GM-SRM} significantly outperforms Lahiri \emph{et al.} on all datasets, which also explicitly learns the generative priors by simply pre-training the decoder. It reveals the effectiveness of the proposed generative memory in \emph{GM-SRM} for learning semantic priors.

Table~\ref{tab:center} lists experimental results for challenging center masked images. Unlike former experiments that only compare 25\% corrupted ratios, we provide results for both 25\% and 50\% corrupted ratios from continuous center masked corruption to evaluate the performance of methods on image inpainting with the small and large corrupted area, respectively. As shown in Table~\ref{tab:center}, \emph{GM-SRM} outperforms other state-of-the-art methods in terms of all five metrics on three benchmark datasets. It demonstrates \emph{GM-SRM} can synthesize more reasonable content than other methods, especially in cases with large corrupted area.


\input{Tables/table3}

\smallskip\noindent\textbf{Qualitative Evaluation.}
We compare the restored results by our \emph{GM-SRM} and other state-of-the-art image inpainting methods on the Paris Street View~\cite{doersch2012makes}, CelebA-HQ~\cite{karras2018progressive}, and Places2~\cite{zhou2017places} in Figure~\ref{fig:case1}, \ref{fig:case2} and \ref{fig:case3}.
PConv~\cite{liu2018image} is specifically proposed to handle irregular corruption, and thus it restores plausible results in Figure~\ref{fig:case1}(b) and Figure~\ref{fig:case2}(b). However, it even cannot fill in reasonable structures when restoring large continuous corruption. DeepFillv2~\cite{yu2019free} normalizes feature maps by the soft mask mechanism, and this relieves the limitation of PConv. Although convincing structures are synthesized in corrupted areas, undesired artifacts can still be observed in Figure~\ref{fig:case2}(c). LISK~\cite{jie2020inpainting} integrates structural information to infer corrupted regions, which may leads to confusing content for the error prediction of image structure(see Figure~\ref{fig:case2})(d). MEDFE~\cite{Liu2019MEDFE} restores structure and texture at the same time. Thus, it predicts stable structure than other competing methods from Figure~\ref{fig:case2}. And yet MEDFE produces unwanted blurriness and color discrepancy(see Figure~\ref{fig:case3}(e)). Conversely, our \emph{GM-SRM} synthesizes higher quality content and more reasonable semantic structures. From Figure~\ref{fig:case1}, \ref{fig:case2} and \ref{fig:case3}, \emph{GM-SRM} avoids undesired blurriness, artifacts and color discrepancy to making restored images more realistic.

Compared with Lahiri \emph{et al.} which also explicitly learns the semantic priors for image inpainting, our \emph{GM-SRM} is able to restore content for the corrupted area that is more realistic in pixel level and more reasonable in semantic level. These qualitative comparisons illustrate the advantages of our model over Lahiri \emph{et al.} in learning the semantic priors.

\smallskip\noindent\textbf{User Study.} 
Quantitative metrics have their bias for the quality evaluation of restored images. To standardize evaluation process, we further perform user study with another four state-of-the-art methods for image inpainting, including DeepFillv2~\cite{yu2019free}, LISK~\cite{jie2020inpainting}, Lahiri \emph{et al.}~\cite{lahiri2020prior} and MEDFE~\cite{Liu2019MEDFE}. We randomly select 50 test images of CelebA-HQ~\cite{karras2018progressive} and Places2~\cite{zhou2017places} respectively, and present restoration results of four methods to 50 human subjects for manual ranking of image quality. Table~\ref{tab:user} lists the voting results of this user study. For the CelebA-HQ dataset, our \emph{GM-SRM} reaches $60.92\%$ votes among total 50$\times$50=2500 rankings, which is much higher than other competing methods. In addition, we count winning samples of each method, and our \emph{GM-SRM} wins on 39 test samples and others altogether 11 samples. As for the Places2 dataset, our \emph{GM-SRM} reaches $58.64\%$ votes among 2500 rankings, which is much higher than other competing methods as well. And our \emph{GM-SRM} wins on 32 test samples and others altogether 18 samples.

\subsection{Investigation on GM-SRM by Ablation Study}
\label{sec:ablation}
We conduct ablation study to investigate the effect of different technical components in \emph{GM-SRM}. To such end, we perform experiments on four variants of our \emph{GM-SRM}.
\begin{itemize}[leftmargin=*]
\setlength{\itemsep}{0pt}

\begin{figure*}[!t]
  \centering 
  \begin{minipage}[b]{0.975\linewidth}
  \includegraphics[width=\linewidth]{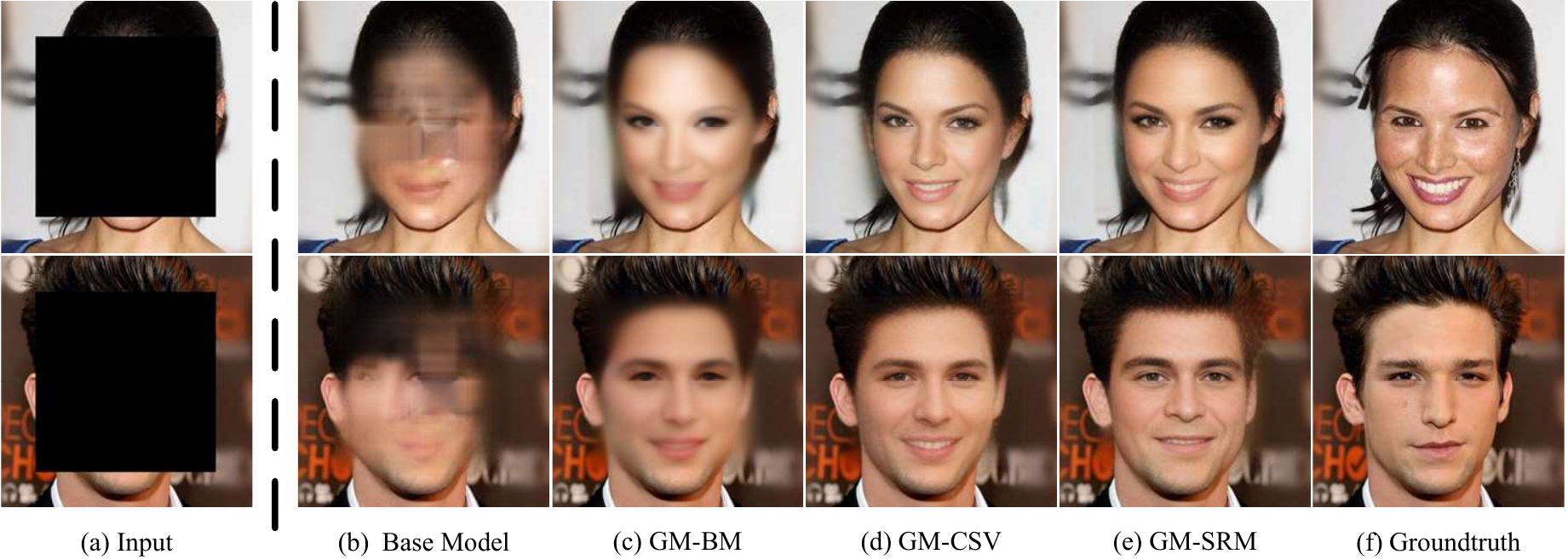}
  \end{minipage}
  \vspace{-8pt}
\caption{Visualization of inpainted images from 50\% center corruption by four variants of the \emph{GM-SRM} for image inpainting on two samples which are randomly selected from CelebA-HQ test set. Best viewed in zoom-in mode.}
\vspace{-4pt}
\label{fig:ablation}
\end{figure*}

\item \smallskip\noindent\textbf{Base model}, which only employs an encoder-decoder model as the base network to restore corrupted images. Thus, no generative memory or stochastic variation strategies are leveraged.


\item \smallskip\noindent\textbf{GM-BM}, which further leverages decoder to integrate features from both generative memory and encoder, and this is equivalent to plug \textbf{G}enerative \textbf{M}emory in \textbf{B}ase \textbf{M}odel as semantic query unit for image generation.

\item \smallskip\noindent\textbf{GM-CSV}, which leverages \textbf{G}enerative \textbf{M}emory with \textbf{C}onditional \textbf{S}tochastic \textbf{V}ariation strategy to obtain more realist details in restored area.

\item \smallskip\noindent\textbf{\emph{GM-SRM}}, which further leverages progressive reasoning strategy to stabilize generated structure and improve image quality.The resulting model is our intact \emph{GM-SRM}.

\end{itemize}
Table~\ref{tab:ablation} illustrates the results of four variants of \emph{GM-SRM} on CelebA-HQ dataset, in terms of PSNR, SSIM, NCC, MSE and LPIPS.

\noindent\textbf{Qualitative ablation evaluation}. As illustrated in Figure~\ref{fig:ablation}, restoration results of four variants of \emph{GM-SRM} are visualized. The results demonstrate that the visual quality of restored content is increasingly better as the augment of \emph{GM-SRM} with different functional components.

\noindent\textbf{Effect of generative memory}. As illustrated in Table~\ref{tab:ablation}, the large gap between \textbf{Base model} and \textbf{GM-BM} demonstrate that our proposed generative memory improves image inpainting performance significantly compared to typical encoder-decoder architecture. Combined pixel-level content reasoning with generative semantic reasoning, generative memory facilitates inpainting reasonable results while coping with large corrupted areas.
\input{Tables/table4}

\noindent\textbf{Effect of conditional stochastic variation}. The conditional stochastic variation improves the texture quality of synthesized images due to learning conditional distribution from known information as a constraint. Intuitive results are shown in Figure~\ref{fig:noise} and Figure~\ref{fig:ablation}.

\noindent\textbf{Effect of progressive reasoning strategy}. Comparison between \textbf{GM-CSV} and \textbf{GM-SRM} demonstrates that progressive reasoning strategy indeed improves the inpainting quality. In Figure~\ref{fig:ablation}, the \emph{GM-SRM} makes the inferred structure and texture more realistic. For instance, in Figure~\ref{fig:ablation}(e), the GM-CSV generates a weird ear, which seems too long, but the \emph{GM-SRM} corrects it. Considering information increment from feature reasoning in different scales, the progressive strategy makes the inpainting results more reasonable according to the known content.




%% file: Tables/table1.tex
\begin{table*}[!t]
\centering
\caption{Performance of different models for irregular image inpainting on the Paris Street View~\cite{doersch2012makes}, CelebA-HQ256~\cite{karras2018progressive}, and Places~\cite{zhou2017places} datasets. Metrics with $\uparrow$, higher value denotes better performance, whereas $\downarrow$ denotes lower is better. 
The best results of each metric are highlighted in bold.}
\resizebox{0.95\linewidth}{!}{
\begin{tabular}{c l| c c c c c c c c c c c c}
\toprule
\multirow{2}*{~}&\multirow{2}*{Method}&\multicolumn{4}{c}{Paris Street View} &\multicolumn{4}{c}{CelebA-HQ}&\multicolumn{4}{c}{Places2} \\
\cmidrule(lr){3-6} 
\cmidrule(lr){7-10}
\cmidrule(lr){11-14}
~& ~ &20-30\% & 30-40\% & 40-50\% & 50-60\%&20-30\% & 30-40\% & 40-50\% & 50-60\%&20-30\% & 30-40\% & 40-50\% & 50-60\% \\
\midrule
\multirow{10}*{\rotatebox{90}{PSNR~$\uparrow$}}&PConv~\cite{
liu2018image}&26.85&24.66&22.67&20.52&28.29&26.11&23.90&21.47&25.05&22.99&21.11&19.04 \\
~&DeepFill V2~\cite{yu2019free}&26.29&23.99&21.86&19.79&27.81&25.52&23.19&20.64&25.15&23.00&20.94&18.68 \\
~&EdgeConnect~\cite{Nazeri_2019_ICCV}&27.62&25.39&23.19&20.86&28.53&26.14&23.53&20.38&25.75&23.64&21.66&19.46 \\
~&LISK~\cite{jie2020inpainting}&25.04&23.10&21.15&19.00&30.22&27.63&24.95&21.51&27.04&24.68&22.35&19.77 \\
~&LBAM~\cite{xie2019image}&27.54&25.25&23.09&20.84&28.86&26.55&24.22&21.68&25.69&23.46&21.41&19.21\\
~&E2I~\cite{xu2020e2i}&26.78&23.80&21.94&19.14&26.28&25.03&23.19&20.20&24.91&23.14&21.09&18.56\\
~&CTSDG~\cite{Guo_2021_ICCV}&27.33&24.67&22.00&19.98&28.49&25.99&23.77&21.22&26.35&24.11&21.58&19.90\\
~&MEDFE~\cite{Liu2019MEDFE}&27.23&24.78&22.30&19.84&25.50&24.10&22.31&20.04&24.84&23.38&21.38&18.27 \\
~&Lahiri \emph{et al.}~\cite{lahiri2020prior}&27.26&24.71&22.76&20.58&27.96&25.89&24.73&21.26&25.54&23.51&21.60&19.41 \\
~&GM-SRM (ours)&\textbf{29.04}&\textbf{26.79}&\textbf{24.51}&\textbf{22.20}&\textbf{30.23}&\textbf{27.91}&\textbf{25.51}&\textbf{22.70}&\textbf{27.09}&\textbf{24.93}&\textbf{22.91}&\textbf{20.60} \\
\midrule
\multirow{10}*{\rotatebox{90}{SSIM~$\uparrow$}}&PConv~\cite{liu2018image}&0.861&0.795&0.711&0.622&0.896&0.847&0.783&0.711&0.844&0.777&0.693&0.611 \\
~&DeepFill V2~\cite{yu2019free}&0.865&0.801&0.714&0.625&0.894&0.842&0.772&0.693&0.859&0.796&0.714&0.630 \\
~&EdgeConnect~\cite{Nazeri_2019_ICCV}&0.879&0.821&0.740&0.651&0.905&0.856&0.784&0.689&0.862&0.799&0.717&0.633 \\
~&LISK~\cite{jie2020inpainting}&0.881&0.823&0.744&0.642&\textbf{0.928}&\textbf{0.886}&0.822&0.724&0.882&0.823&0.744&0.655 \\
~&LBAM~\cite{xie2019image}&0.879&0.816&0.732&0.637&0.907&0.860&0.796&0.718&0.861&0.794&0.708&0.619 \\
~&E2I~\cite{xu2020e2i}&0.865&0.810&0.718&0.620&0.895&0.846&0.779&0.698&0.861&0.798&0.719&0.634\\
~&CTSDG~\cite{Guo_2021_ICCV}&0.874&0.819&0.728&0.629&0.905&0.889&0.780&0.697&0.869&0.814&0.728&0.649\\
~&MEDFE~\cite{Liu2019MEDFE}&0.876&0.811&0.718&0.616&0.884&0.810&0.759&0.666&0.833&0.809&0.738&0.636 \\
~&Lahiri \emph{et al.}~\cite{lahiri2020prior}&0.871&0.813&0.739&0.651&0.903&0.859&0.802&0.716&0.869&0.811&0.738&0.666 \\
~&GM-SRM (ours)&\textbf{0.902}&\textbf{0.852}&\textbf{0.780}&\textbf{0.696}&0.925&\textbf{0.886}&\textbf{0.832}&\textbf{0.760}&\textbf{0.883}&\textbf{0.827}&\textbf{0.756}&\textbf{0.677} \\
\midrule
\multirow{10}*{\rotatebox{90}{NCC~$\uparrow$}}&PConv~\cite{liu2018image}&0.957&0.934&0.895&0.835&0.986&0.978&0.964&0.938&0.957&0.934&0.898&0.838 \\
~&DeepFill V2~\cite{yu2019free}&0.951&0.924&0.878&0.811&0.985&0.975&0.958&0.926&0.956&0.933&0.894&0.826 \\
~&EdgeConnect~\cite{Nazeri_2019_ICCV}&0.963&0.943&0.906&0.848&0.987&0.978&0.960&0.920&0.962&0.942&0.908&0.850 \\
~&LISK~\cite{jie2020inpainting}&0.971&0.954&0.919&0.859&\textbf{0.991}&\textbf{0.985}&0.972&0.940&\textbf{0.972}&0.954&0.921&0.862 \\
~&LBAM~\cite{xie2019image}&0.962&0.941&0.904&0.845&0.987&0.980&0.966&0.940&0.962&0.941&0.906&0.848 \\
~&E2I~\cite{xu2020e2i}&0.956&0.929&0.881&0.820&0.986&0.979&0.961&0.928&0.959&0.936&0.896&0.830\\
~&CTSDG~\cite{Guo_2021_ICCV}&0.959&0.940&0.900&0.822&0.982&0.979&0.959&0.931&0.960&0.947&0.913&0.835\\
~&MEDFE~\cite{Liu2019MEDFE}&0.961&0.937&0.889&0.814&0.976&0.967&0.950&0.917&0.956&0.939&0.906&0.813 \\
~&Lahiri \emph{et al.}~\cite{lahiri2020prior}&0.964&0.947&0.916&0.854&0.988&0.981&0.969&0.937&0.968&0.952&0.925&0.876 \\
~&GM-SRM (ours)&\textbf{0.972}&\textbf{0.958}&\textbf{0.929}&\textbf{0.879}&\textbf{0.991}&\textbf{0.985}&\textbf{0.974}&\textbf{0.952}&\textbf{0.972}&\textbf{0.956}&\textbf{0.929}&\textbf{0.880} \\
\midrule
\multirow{10}*{\rotatebox{90}{LMSE~$\downarrow$}}&PConv~\cite{liu2018image}&0.016&0.025&0.037&0.051&0.007&0.011&0.017&0.026&0.018&0.027&0.038&0.053 \\
~&DeepFill V2~\cite{yu2019free}&0.019&0.030&0.046&0.064&0.008&0.013&0.020&0.031&0.019&0.029&0.043&0.062 \\
~&EdgeConnect~\cite{Nazeri_2019_ICCV}&0.013&0.020&0.044&0.064&0.006&0.010&0.017&0.031&0.015&0.023&0.034&0.048 \\
~&LISK~\cite{jie2020inpainting}&0.010&0.016&0.027&0.040&\textbf{0.004}&0.007&0.013&0.024&\textbf{0.011}&0.019&0.030&0.046 \\
~&LBAM~\cite{xie2019image}&0.014&0.022&0.033&0.047&0.006&0.010&0.016&0.025&0.016&0.025&0.037&0.053 \\
~&E2I~\cite{xu2020e2i}&0.015&0.027&0.031&0.052&0.008&0.012&0.018&0.029&0.016&0.027&0.039&0.051\\
~&CTSDG~\cite{Guo_2021_ICCV}&0.012&0.022&0.035&0.048&0.009&0.010&0.019&0.028&0.015&0.024&0.035&0.052\\
~&MEDFE~\cite{Liu2019MEDFE}&0.014&0.023&0.038&0.055&0.011&0.015&0.021&0.032&0.018&0.025&0.037&0.058 \\
~&Lahiri \emph{et al.}~\cite{lahiri2020prior}&0.011&0.020&0.029&0.049&0.006&0.009&0.013&0.024&0.014&0.019&0.040&0.046 \\
~&GM-SRM (ours)&\textbf{0.009}&\textbf{0.015}&\textbf{0.023}&\textbf{0.034}&\textbf{0.004}&\textbf{0.006}&\textbf{0.011}&\textbf{0.018}&\textbf{0.011}&\textbf{0.017}&\textbf{0.025}&\textbf{0.037} \\
\midrule
\multirow{10}*{\rotatebox{90}{LPIPS~$\downarrow$}}&PConv~\cite{liu2018image}&0.123&0.162&0.208&0.282&0.047&0.078&0.100&0.157&0.098&0.140&0.202&0.279 \\
~&DeepFill V2~\cite{yu2019free}&0.118&0.154&0.191&0.269&0.054&0.080&0.117&0.164&0.079&0.119&0.178&0.256 \\
~&EdgeConnect~\cite{Nazeri_2019_ICCV}&0.081&0.114&0.174&0.210&0.045&0.067&0.106&0.165&0.075&0.107&0.164&0.279 \\
~&LISK~\cite{jie2020inpainting}&0.072&\textbf{0.098}&0.168&0.209&0.039&0.051&0.082&0.134&0.076&0.110&\textbf{0.161}&0.247 \\
~&LBAM~\cite{xie2019image}&0.069&0.105&0.143&0.257&0.038&0.057&0.087&\textbf{0.128}&0.074&0.113&0.172&0.250 \\
~&E2I~\cite{xu2020e2i}&0.094&0.137&0.188&0.260&0.049&0.072&0.107&0.160&0.077&0.111&0.178&0.261 \\
~&CTSDG~\cite{Guo_2021_ICCV}&0.074&0.120&0.171&0.245&0.040&0.068&0.094&0.151&0.080&0.122&0.188&0.253\\
~&MEDFE~\cite{Liu2019MEDFE}&0.101&0.127&0.176&0.238&0.052&0.084&0.115&0.166&0.105&0.130&0.190&0.290 \\
~&Lahiri \emph{et al.}~\cite{lahiri2020prior}&0.095&0.114&0.165&0.286&0.046&0.069&0.101&0.149&0.080&0.124&0.180&0.259 \\
~&GM-SRM (ours)&\textbf{0.063}&\textbf{0.098}&\textbf{0.138}&\textbf{0.207}&\textbf{0.031}&\textbf{0.048}&\textbf{0.079}&\textbf{0.128}&\textbf{0.071}&\textbf{0.109}&\textbf{0.161}&\textbf{0.234} \\

\bottomrule
\end{tabular}
 }
\label{tab:irregular}
\vspace{-4pt}
\end{table*}

%% file: Tables/table2.tex
\begin{table}[!t]
\centering
\caption{Performance of different models for regular image inpainting on the Paris Street View~\cite{doersch2012makes}, CelebA-HQ256~\cite{karras2018progressive}, and Places2~\cite{zhou2017places} datasets. Metrics with $\uparrow$, higher value denotes better performance, whereas $\downarrow$ denotes lower is better. 
The best results of each metric are highlighted in bold.}
\resizebox{0.95\linewidth}{!}{
\begin{tabular}{c l| c c c c c c}
\toprule
\multirow{2}*{~}&\multirow{2}*{Method}&\multicolumn{2}{c}{Paris Street View} &\multicolumn{2}{c}{CelebA-HQ}&\multicolumn{2}{c}{Places2} \\
\cmidrule(lr){3-4} 
\cmidrule(lr){5-6}
\cmidrule(lr){7-8}
~& ~&25\% & 50\% &25\% & 50\% &25\% & 50\%  \\
\midrule
\multirow{10}*{\rotatebox{90}{PSNR~$\uparrow$}}&PConv~\cite{liu2018image}&23.97&20.02&20.54&14.56&21.69&18.09 \\
~&DeepFill V2~\cite{yu2019free}&23.32&18.98&25.45&19.86&21.27&17.54 \\
~&EdgeConnect~\cite{Nazeri_2019_ICCV}&24.91&20.32&23.68&18.83&22.39&18.48 \\
~&LISK~\cite{jie2020inpainting}&22.28&18.32&24.28&18.02&22.52&18.54 \\
~&LBAM~\cite{xie2019image}&24.18&20.14&25.91&20.73&21.79&18.38 \\
~&E2I~\cite{xu2020e2i}&23.40&19.27&25.59&19.99&21.32&18.17\\
~&CTSDG~\cite{Guo_2021_ICCV}&24.22&19.67&25.51&20.08&22.27&18.40\\
~&MEDFE~\cite{Liu2019MEDFE}&23.43&19.09&25.54&20.26&21.18&17.57\\
~&Lahiri \emph{et al.}~\cite{lahiri2020prior}&24.79&19.92&25.51&20.50&22.35&18.36 \\
~&GM-SRM(ours)&\textbf{25.86}&\textbf{21.41}&\textbf{26.97}&\textbf{21.75}&\textbf{23.33}&\textbf{19.51} \\
\midrule
\multirow{10}*{\rotatebox{90}{SSIM~$\uparrow$}}&PConv~\cite{liu2018image}&0.828&0.638&0.797&0.596&0.816&0.633 \\
~&DeepFill V2~\cite{yu2019free}&0.832&0.633&0.876&0.694&0.823&0.638 \\
~&EdgeConnect~\cite{Nazeri_2019_ICCV}&0.846&0.659&0.860&0.672&0.827&0.647 \\
~&LISK~\cite{jie2020inpainting}&0.828&0.639&0.873&0.675&0.842&0.662 \\
~&LBAM~\cite{xie2019image}&0.832&0.637&0.887&0.719&0.820&0.634 \\
~&E2I~\cite{xu2020e2i}&0.834&0.636&0.879&0.699&0.825&0.640\\
~&CTSDG~\cite{Guo_2021_ICCV}&0.831&0.638&0.874&0.691&0.828&0.641\\
~&MEDFE~\cite{Liu2019MEDFE}&0.827&0.615&0.880&0.703&0.813&0.622 \\
~&Lahiri \emph{et al.}~\cite{lahiri2020prior}&0.835&0.649&0.882&0.701&0.830&0.669 \\
~&GM-SRM(ours)&\textbf{0.862}&\textbf{0.695}&\textbf{0.901}&\textbf{0.757}&\textbf{0.845}&\textbf{0.683} \\
\midrule
\multirow{10}*{\rotatebox{90}{NCC~$\uparrow$}}&PConv~\cite{liu2018image}&0.924&0.817&0.894&0.691&0.909&0.800 \\
~&DeepFill V2~\cite{yu2019free}&0.913&0.781&0.974&0.911&0.899&0.781 \\
~&EdgeConnect~\cite{Nazeri_2019_ICCV}&0.937&0.829&0.962&0.887&0.920&0.815 \\
~&LISK~\cite{jie2020inpainting}&0.939&0.833&0.968&0.871&0.930&0.833 \\
~&LBAM~\cite{xie2019image}&0.928&0.820&0.976&0.926&0.917&0.820 \\
~&E2I~\cite{xu2020e2i}&0.918&0.794&0.977&0.918&0.903&0.792\\
~&CTSDG~\cite{Guo_2021_ICCV}&0.925&0.811&0.969&0.911&0.916&0.818\\
~&MEDFE~\cite{Liu2019MEDFE}&0.919&0.782&0.974&0.918&0.902&0.781 \\
~&Lahiri \emph{et al.}~\cite{lahiri2020prior}&0.929&0.830&0.969&0.926&0.919&0.830 \\
~&GM-SRM(ours)&\textbf{0.946}&\textbf{0.860}&\textbf{0.981}&\textbf{0.940}&\textbf{0.932}&\textbf{0.840} \\
\midrule
\multirow{10}*{\rotatebox{90}{LMSE~$\downarrow$}}&PConv~\cite{liu2018image}&0.024&0.051&0.023&0.044&0.027&0.054 \\
~&DeepFill V2~\cite{yu2019free}&0.029&0.069&0.012&0.034&0.034&0.068 \\
~&EdgeConnect~\cite{Nazeri_2019_ICCV}&0.019&0.046&0.017&0.038&0.024&0.051 \\
~&LISK~\cite{jie2020inpainting}&0.017&0.044&0.015&0.032&0.020&0.047 \\
~&LBAM~\cite{xie2019image}&0.022&0.049&0.011&0.028&0.028&0.055 \\
~&E2I~\cite{xu2020e2i}&0.025&0.057&0.012&0.033&0.031&0.056\\
~&CTSDG~\cite{Guo_2021_ICCV}&0.024&0.050&0.013&0.032&0.025&0.052\\
~&MEDFE~\cite{Liu2019MEDFE}&0.022&0.060&0.012&0.033&0.030&0.063 \\
~&Lahiri \emph{et al.}~\cite{lahiri2020prior}&0.026&0.047&0.020&0.034&0.030&0.050 \\
~&GM-SRM(ours)&\textbf{0.015}&\textbf{0.035}&\textbf{0.008}&\textbf{0.022}&\textbf{0.017}&\textbf{0.038} \\
\midrule
\multirow{10}*{\rotatebox{90}{LPIPS~$\downarrow$}}&PConv~\cite{liu2018image}&0.110&0.242&0.079&0.341&0.146&0.301 \\
~&DeepFill V2~\cite{yu2019free}&0.106&0.239&0.056&0.166&0.133&0.277 \\
~&EdgeConnect~\cite{Nazeri_2019_ICCV}&0.099&0.219&0.072&0.185&0.122&\textbf{0.263} \\
~&LISK~\cite{jie2020inpainting}&0.096&0.226&0.067&0.298&0.149&0.287 \\
~&LBAM~\cite{xie2019image}&0.100&0.222&0.047&0.135&0.135&0.272 \\
~&E2I~\cite{xu2020e2i}&0.105&0.233&0.056&0.159&0.131&0.270 \\
~&CTSDG~\cite{Guo_2021_ICCV}&0.103&0.225&0.059&0.160&0.142&0.281\\
~&MEDFE~\cite{Liu2019MEDFE}&0.114&0.250&0.052&0.143&0.160&0.312 \\
~&Lahiri \emph{et al.}~\cite{lahiri2020prior}&0.117&0.277&0.053&0.154&0.151&0.325 \\
~&GM-SRM (ours)&\textbf{0.093}&\textbf{0.218}&\textbf{0.045}&\textbf{0.132}&\textbf{0.120}&\textbf{0.263} \\
\bottomrule
\end{tabular}}
\label{tab:center}
\vspace{-4pt}
\end{table}

%% file: Tables/table3.tex
\begin{table}[!t]
\centering
\caption{User study on 100 restored results of Places2~\cite{zhou2017places}, and CelebA-HQ256~\cite{karras2018progressive}. 50 human subjects are performed for comparison between our \emph{GM-SRM} and state-of-the-art methods. 
}
\resizebox{0.9\linewidth}{!}{
\begin{tabular}{l| c c c c}
\toprule
\multirow{2}*{Method} & \multicolumn{2}{c}{Share of the vote} & \multicolumn{2}{c}{Winning samples}\\
\cmidrule(lr){2-3} 
\cmidrule(lr){4-5}
~ & CelebA-HQ & Places2 & CelebA-HQ & Places2 \\
\midrule
DeepFill V2~\cite{yu2019free}&3.96\%&8.92\%&1&3\\
LISK~\cite{jie2020inpainting}&15.84\%&16.60\%&4&6\\
Lahiri \emph{et al.}~\cite{lahiri2020prior}&0.28\%&4.68\%&0&2\\
MEDFE~\cite{Liu2019MEDFE}&19.00\%&11.16\%&6&7\\
GM-SRM&60.92\%&58.64\%&39&32 \\
\bottomrule
\end{tabular}
}
\label{tab:user}
\vspace{-8pt}
\end{table} 

%% file: Tables/table4.tex
\begin{table}[!t]
\centering
\caption{Ablation experiments of 50\% corruption ratio on CelebA-HQ~\cite{karras2018progressive} in terms of PSNR, SSIM, NCC, LMSE, and LPIPS to investigate effect of each component in our \emph{GM-SRM}.}
\resizebox{0.75\linewidth}{!}{
\begin{tabular}{l| c c c c c}
\toprule
Method&PSNR&SSIM&NCC&LMSE&LPIPS\\
\midrule
Base model&18.39&0.694&0.874&0.032&0.235\\
GM-BM&20.43&0.729&0.921&0.027&0.143\\
GM-CSV&21.06&0.734&0.930&0.023&0.139\\
\textbf{GM-SRM}&\textbf{21.75}&\textbf{0.757}&\textbf{0.940}&\textbf{0.022}&\textbf{0.132}\\
\bottomrule
\end{tabular}
}
\label{tab:ablation}
\end{table} 

%% file: 5.Conclusion.tex
In this work, we have presented the Generative Memory-guided Semantic Reasoning Model (\emph{GM-SRM}) for image inpainting, which infers the corrupted content based on not only the known regions of the input image, but also the explicitly learned inter-image reasoning priors characterizing the generalizable semantic distribution patterns between similar images.
Our \emph{GM-SRM} employ the encoder-decoder framework to guarantee the pixel-level content consistency, and our pre-trained generative memory is favorable for performing high-level semantic reasoning. 
Compared to previous implicitly learned semantic priors, our generative priors favor high-level semantic reasoning. 
As a result, our model is able to synthesize semantically reasonable content for the corrupted regions, especially in the scenarios with large corrupted area.